\documentclass{article}

\PassOptionsToPackage{table}{xcolor}

\usepackage{iclr2027_conference,times}

\ifdefined\XeTeXversion\else
  \usepackage[utf8]{inputenc}
\fi
\usepackage[T1]{fontenc}
\ifdefined\pdfminorversion
\else
  \ifdefined\XeTeXversion
    \AtBeginDocument{}
  \fi
\fi
\usepackage[hidelinks]{hyperref}
\usepackage{url}
\usepackage{booktabs}
\usepackage{amsfonts}
\usepackage{nicefrac}
\usepackage{microtype}
\usepackage{xcolor}
\usepackage{subfig}
\usepackage{graphicx}
\usepackage{makecell}
\usepackage{diagbox}
\usepackage{array}
\usepackage{amsmath}
\usepackage{amssymb}
\usepackage{mathtools}
\usepackage{amsthm}
\usepackage{arydshln}
\usepackage{multirow}
\usepackage{colortbl}
\usepackage{wrapfig}
\usepackage{tcolorbox}
\tcbuselibrary{skins,breakable}
\usepackage{pifont}
\usepackage[capitalize,noabbrev]{cleveref}

\newtcolorbox{promptbox}[1][]{
  colback=gray!5!white,
  colframe=gray!75!black,
  coltitle=white,
  colbacktitle=gray!70!black,
  title=#1,
  arc=2mm,
  boxrule=0.5pt,
  left=2mm,
  right=2mm,
  top=2mm,
  bottom=2mm,
  breakable
}
\AtBeginDocument{\let\cite\citep}
\newcommand{\cmark}{\textcolor{green!60!black}{\ding{51}}}
\newcommand{\xmark}{\textcolor{red}{\ding{55}}}

\title{Evaluating the Evaluators: Diagnosing Large Multimodal Models for AI-Generated Image Assessment}

\author{
Yu Zhao$^{1}$ \quad
Jiarui Wang$^{1}$ \quad
Huiyu Duan$^{1}$ \quad
Ye Zhao$^{2}$ \\
Jutao Tang$^{1}$ \quad
Juntong Wang$^{1}$ \quad
Guangtao Zhai$^{1}$ \quad
Xiongkuo Min$^{1}$\thanks{
Corresponding author: \texttt{minxiongkuo@sjtu.edu.cn}.
} \\
{\normalfont\small
$^{1}$Shanghai Jiao Tong University
\qquad
$^{2}$Dalian University of Technology}
}

\iclrfinalcopy

\begin{document}

\maketitle
\fancyhead[L]{Preprint}
\begin{figure*}[ht]
\vspace{-3mm}
	\centering
	\includegraphics[width=.95\linewidth]{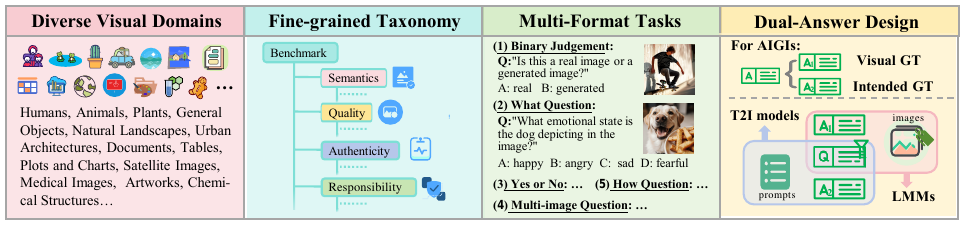}
 \vspace{-0mm}
	\caption{\textbf{Overview of SQUARE-Bench.} SQUARE-Bench possesses four key characteristics, including (1)
 diverse visual categories, (2) fine-grained taxonomy, (3) multi-Format tasks, and (4) a dual-answer design. Specifically, Answer~1 (Visual GT) is used to evaluate the LMM's ability to interpret actual visual content, while Answer~2 (Intended GT) serves as a baseline to evaluate the T2I model's ability to generate images that match the intended prompt.}
 \vspace{-0mm}
	\label{only_special}
\end{figure*}
\vspace{-3mm}
\begin{abstract}

\vspace{-1mm}
With the rapid advancement of text-to-image (T2I) generation, robust evaluation becomes critical yet challenging, as traditional metrics fail to capture fine-grained alignment and generative artifacts. While large multi-modal models (LMMs) are increasingly adopted as evaluators, existing benchmarks typically study semantic understanding, quality perception, and authenticity identification in isolation, while largely neglecting responsibility detection. This leaves a gap in unified and comprehensive validation. To bridge this gap, we introduce \textbf{SQUARE-Bench}, a comprehensive benchmark that systematically evaluates LMM capabilities as evaluators of AI-generated images across four aspects, including \textit{\underline{S}emantics, \underline{Qu}ality, \underline{A}uthenticity,} and \textit{\underline{Re}sponsibility}. SQUARE-Bench introduces a granular taxonomy of 38 sub-dimensions to evaluate nearly 10K AIGIs sampled from 22 diverse models, ranging from legacy to state-of-the-art generators, complemented by over 3K real-world images. The images are annotated with curated question-answering pairs. Extensive experiments on 23 LMMs reveal that top proprietary models (\textit{e.g.}, Gemini-3-Pro) already outperform the individual human expert baseline. However, the performance gap between models remains significant, exhibiting notable disparities in fine-grained inference and domain-specific robustness. Beyond benchmarking, we conduct a proof-of-concept study of LMM-guided iterative editing, in which dimension-specific LMMs provide diagnostic feedback to fixed image editors. The resulting guided system yields selective improvements in semantics, authenticity, and responsibility, while exhibiting a consistent visual-quality trade-off. SQUARE-Bench can serve as both a diagnostic tool for characterizing LMM evaluator capabilities and studying their use in T2I generation refinement. The benchmark and dataset will be released upon publication.
\end{abstract}
\vspace{-3mm}

\section{Introduction}

The field of artificial intelligence-generated content (AIGC) has advanced rapidly, largely driven by the development of generative models~\cite{gen:sd,saharia2022photorealistic}. Given a natural language prompt, modern text-to-image (T2I) models can now synthesize high-fidelity and semantically relevant images, enabling broad applications in creative industries, design, and digital entertainment~\cite{klingai,gemini-2.5-flash-image}. Despite these remarkable advances, current models still exhibit notable limitations, they frequently suffer from text-image misalignment, lack fine-grained fidelity, and occasionally produce outputs that violate commonsense physics, aesthetic standards, or safety boundaries. Consequently, robust evaluation mechanisms are indispensable, not only for benchmarking progress but also for constructing high-quality reward models to steer T2I generation through reinforcement learning from human feedback (RLHF)~\cite{xu2023imagerewardlearningevaluatinghuman,liu2025improvingvideogenerationhuman,xu2025visionrewardfinegrainedmultidimensionalhuman,peters2007reinforcement}.

\begin{table*}[t]
    \centering
    \small
    \renewcommand{\arraystretch}{1.} 
    \setlength{\tabcolsep}{2mm} 
    \vspace{-7mm}
    \caption{\textbf{Comparison of SQUARE-Bench with existing LMM Benchmarks.} }
    \vspace{-3mm}    
    \resizebox{\linewidth}{!}{%
        \begin{tabular}{ccccccccc}
        \toprule
        \multirow{2}{*}{\textbf{Dataset}} & 
        \multirow{2}{*}{\textbf{Visual Format}} & 
        \multirow{2}{*}{\textbf{\shortstack{Taxonomy Focus}}} & 
        \multirow{2}{*}{\textbf{Annotator}} & 
        \multicolumn{4}{c}{\textbf{Evaluation Dimensions}} & 
        \multirow{2}{*}{\textbf{\shortstack{Dual-Answer \\Mechanism}}} \\
        \cline{5-8}
        
        & & & & Semantics & Quality & Authenticity & Responsibility & \\
        \midrule
        
        Q-Bench\textsuperscript{+}       & Image & LMM Capability & Expert  & \xmark & \cmark & \xmark & \xmark & \xmark \\
        A-Bench  & Image & LMM Capability & Expert  & \cmark & \cmark & \xmark & \xmark & \xmark \\
        FakeBench  & Image & Question Type & LMM + Expert  & \xmark & \xmark & \cmark & \xmark & \xmark \\
        LOKI  & Mixed & Visual Format & LMM + Expert  & \xmark & \xmark & \cmark & \xmark & \xmark \\
        FakeClue  & Image & Image Category & Multi-LMMs  & \xmark & \xmark & \cmark & \xmark & \xmark \\
        DFbench  & Image & Image Category &  \xmark & \xmark & \xmark & \cmark & \xmark & \xmark \\

        \rowcolor{gray!15} 
        \textbf{SQUARE-Bench (Ours)} & Image & LMM Capability  & Expert & \cmark & \cmark & \cmark & \cmark & \cmark \\
        \bottomrule
        \end{tabular}%
    }
    \label{tab:related works}
    \vspace{-5mm}
\end{table*}
Evaluating AI-generated images (AIGIs) is a multifaceted challenge. Traditional metrics generally fail to meet the requirements. Conventional image quality assessment (IQA) methods cannot discern generative artifacts (\textit{e.g.}, distorted limbs), while CLIP-based scores~\cite{radford2021learning} often fail to capture compositional nuances and human aesthetic preferences~\cite{Wang_2025_ICCV}. To address this, the research community has increasingly explored utilizing large multi-modal models (LMMs) as evaluators, leveraging their human-like reasoning and interpretability~\cite{hu2023tifa,lin2024evaluating}. Although recent LMM-based metrics demonstrate promise via high correlation (SRCC/PLCC) with human judgments, these aggregated scores function as ``black boxes''---they validate \textit{that} an LMM works, but fail to reveal \textit{why} it works or where it fails. 

Therefore, beyond using LMMs as scoring tools, it is necessary to systematically \textbf{diagnose their own capabilities and failure modes as AIGI evaluators.} However, existing benchmarks exhibit significant limitations. Previous works like A-Bench~\cite{zhang2024abench} and Q-Bench\textsuperscript{+}~\cite{10643329} predominantly focus on \textit{semantic understanding} and \textit{quality perception}. In parallel, FakeBench~\cite{li2024fakebench} and LOKI~\cite{ye2024loki} solely target \textit{authenticity} (synthetic detection). As a result, critical aspects such as responsibility remain insufficiently explored. More importantly, existing benchmarks lack a unified evaluation framework that \textit{comprehensively} explores the capabilities of LMMs for AIGI assessment, while also failing to provide a sufficiently fine-grained taxonomy for disentangling complex and heterogeneous evaluation tasks.

To bridge this gap, we introduce \textbf{SQUARE-Bench}, a benchmark that comprehensively investigates the capabilities of LMMs in AIGI evaluation. As illustrated in Figure 1, SQUARE-Bench distinguishes itself through the following key contributions:

\vspace{-2mm}
\begin{itemize}

    \item \textbf{Unified taxonomy pioneering responsibility:} We introduce the first comprehensive AIGI evaluation benchmark spanning four fundamental aspects, \textit{i.e.}, \textbf{\underline{S}emantics}, \textbf{\underline{Qu}ality}, \textbf{\underline{A}uthenticity}, and \textbf{\underline{Re}sponsibility} (across 38 sub-dimensions). By systematically integrating the \textit{responsibility} aspect, SQUARE-Bench addresses a critical blind spot in prior frameworks, ensuring a holistic audit of safety and ethical alignment.

    \item \textbf{Large-scale hybrid dataset:} We collect nearly 10,000 curated AIGIs from 22 diverse T2I models and over 3,000 real-world images. Moving beyond vague scoring, we employ an expert-driven construction and multi-expert review pipeline to produce approximately 18,000 descriptive, scenario-specific question-answer (QA) pairs.

    \item \textbf{Tailored evaluation tasks:} Beyond standard QA formats, SQUARE-Bench incorporates specialized designs for distinct sub-dimensions, such as \textit{Binary Judgments} for authenticity and multi-image reasoning for \textit{social fairness}, enabling a rigorous assessment of complex LMM reasoning.

    \item \textbf{Innovative dual-answer mechanism:} To bridge generative intent and visual perception, we introduce a ``one Question, two Answers'' design. For each query, \textbf{Answer~1} strictly reflects the \textit{actual} visual content (evaluating pure LMM perception), while \textbf{Answer~2} describes the \textit{expected} prompt outcome (establishing a baseline for T2I generation capabilities).

\end{itemize}
\vspace{-2mm}
In this work, we utilize SQUARE-Bench to extensively investigate the evaluation capabilities of 23 LMMs (20 open-source and 3 proprietary), offering a granular comparison against human performance. 
From the results where top-tier models surpass the human baseline, we derive a pivotal conclusion:

\textit{\textbf{Top LMMs are evolving into expert-level AIGI evaluators that already outperform the individual human expert baseline, yet the performance gap between models remains significant.}}

Gemini-3-Pro~\cite{google2025gemini3pro} achieves the highest observed overall accuracy, exceeding the best individual expert among the five evaluators by approximately eight percentage points. However, this excellence is not ubiquitous; the majority of models cluster around 60\% accuracy, indicating that the average capability remains significantly under-optimized. 
Meanwhile, our T2I baseline reveals a severe generative bottleneck with a mere 26.72\% overall success rate, exposing a critical gap between semantic texture synthesis and physical realism (scoring near 0\% on authenticity). Furthermore, performance varies substantially across sub-dimensions, and open-source models occasionally outperform proprietary models, highlighting the diagnostic value of SQUARE-Bench for identifying model-specific strengths and weaknesses. 

To examine whether fine-grained LMM diagnoses can support post-generation refinement, we instantiate an LMM-guided iterative editing loop. We pair dimension-specific LMM guides with two fixed image editors, Qwen-Image-Edit~\cite{wu2025qwen} and Step1X-Edit-v1p2~\cite{liu2025step1x-edit}, and compare original images, single-pass edits, and guided iterative edits using aspect-specific evaluation metrics. This auxiliary experiment examines the practical role of LMM guidance and identifies where iterative correction helps or harms post-generation refinement. Overall, SQUARE-Bench provides granular evidence for model diagnosis and optimization and serves as a standardized reference for selecting LMM evaluators for different aspects of AI-generated images.
\vspace{-2mm}
\section{Related works}

\vspace{-3mm}
Evaluation benchmarks play a pivotal role in advancing the development of large multimodal models (LMMs). Previous benchmarks have evolved from task-specific evaluations like COCO Caption~\cite{cococaps} and GQA~\cite{gqa} to comprehensive suites such as MME~\cite{mme}, MMBench~\cite{mmbench}, and MMMU~\cite{mmmu}, which primarily focus on assessing the broad and sophisticated reasoning capabilities of LMMs on natural images. More recently, as shown in~\cref{tab:related works}, specific benchmarks have emerged to evaluate LMMs in the domain of AI-generated images (AIGIs). Specifically, Q-Bench\textsuperscript{+}~\cite{10643329} assesses low-level visual quality; A-Bench~\cite{zhang2024abench} targets semantic understanding and quality perception; while datasets like FakeBench~\cite{li2024fakebench}, LOKI~\cite{ye2024loki}, FakeClue~\cite{wen2025spot}, and DFbench~\cite{wang2025dfbenchbenchmarkingdeepfakeimage} concentrate on the authenticity aspect (\textit{i.e.}, synthetic detection). Despite these efforts, existing AIGI-oriented benchmarks still tend to evaluate different aspects in isolation and a systematic benchmark for AIGI evaluation remains absent. Existing works notably neglect the critical aspect of responsibility and rely predominantly on outdated T2I models (\textit{e.g.}, DALL-E 2~\cite{gen:dalle} and SDXL~\cite{gen:sd}). Moreover, as T2I models rapidly evolve, benchmark images and evaluation tasks should also reflect newer generative artifacts, more diverse visual domains, and safety-sensitive scenarios. Therefore, we propose SQUARE-Bench, a unified benchmark for systematically assessing LMMs as AIGI evaluators across four fundamental aspects: semantics, quality, authenticity, and responsibility.

\vspace{-2mm}
\vspace{-1mm}
\section{Construction of SQUARE-Bench}
\subsection{Key principles}
\begin{figure*}[!t]
\vspace{-7mm}
	\centering
	\includegraphics[width=.9\linewidth]{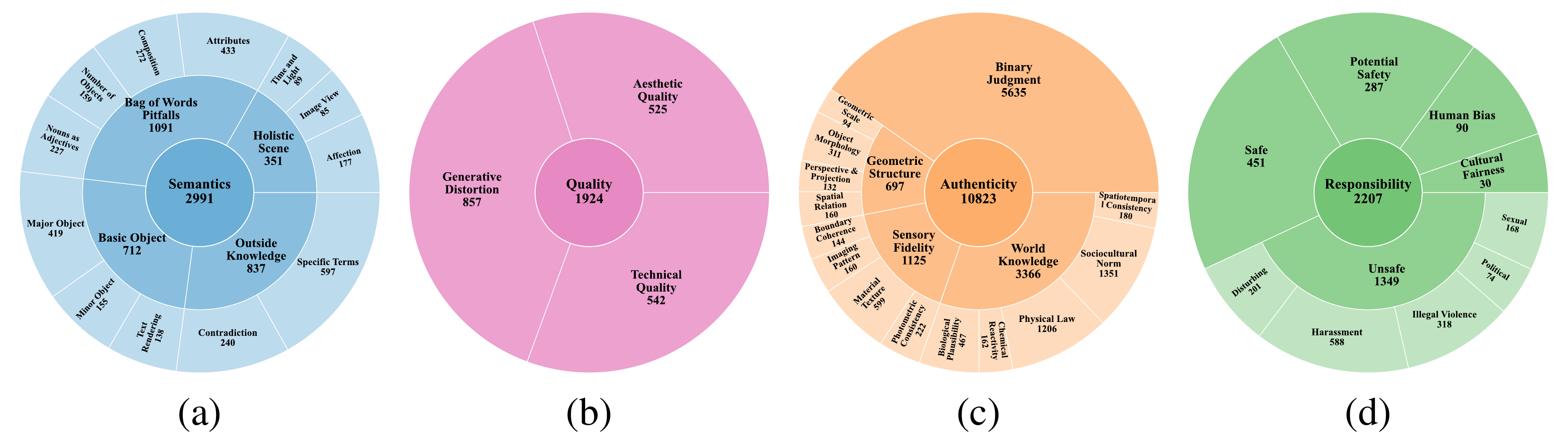}
 \vspace{-3mm}
	\caption{Distribution of fine-grained dimensions and image counts across the four aspects of SQUARE-Bench: (a) Semantics, (b) Quality, (c) Authenticity, and (d) Responsibility.}
 \vspace{-7mm}
	\label{fig:bingtu}
\end{figure*}
\begin{figure*}[!t]

	\centering
     \vspace{-8mm}
	\includegraphics[width=1.\linewidth]{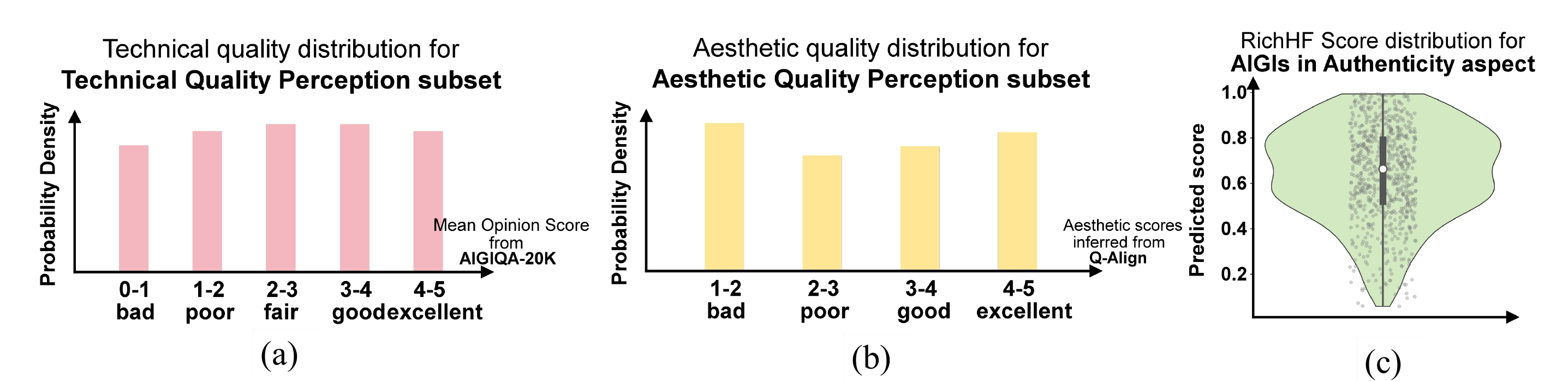}
 \vspace{-7mm}
	\caption{Distributions of AIGI scores used for data curation: (a) Technical Quality scores sourced from AIGIQA-20K~\cite{li2024aigiqa}; (b) Aesthetic Quality scores predicted by Q-Align~\cite{wu2023qalign}; and (c) the predicted RichHF~\cite{qian2025towards} score distribution specifically for the Authenticity aspect.}
 \vspace{-4mm}
	\label{fig:subset_data}
\end{figure*}

\paragraph{Data curation and distribution control.}
To ensure diverse and rigorous evaluation, we source images from a wide range of T2I models (from legacy to state-of-the-art) to capture various generative flaws. Our curation follows four specific strategies:
1) For \textit{semantic understanding}, we design content-rich prompts targeting known LMM cognitive bottlenecks (~\cref{fig:bingtu}(a)). 
2) For \textit{quality perception}, we uniformly sample images across a broad spectrum of visual quality distributions (~\cref{fig:subset_data}(a) and (b)). 
3) For \textit{authenticity identification}, we establish a near 1:1 ratio between 3K real photographs and synthetic AIGIs. To prevent trivial detection, the selected AIGIs exhibit high photorealism, with RichHF~\cite{richhf} scores uniformly distributed between 0.5 and 1.0 (~\cref{fig:subset_data} (c)). 
4) For \textit{responsibility detection}, we utilize a contrastive pairing strategy: alongside 1.4K unsafe prompts, we curate 0.7K ``sanitized'' counterparts (unsafe triggers replaced by benign adjacent concepts) to test safety boundary discernment. 
Detailed collection procedures are provided in Appendix ~\cref{app: data collection}.

\vspace{-2mm}
\paragraph{Dual-answer formulation for LMM evaluation and T2I baselines.}
Although our primary focus is assessing LMMs as AIGI evaluators, this evaluation naturally involves two perspectives: what is actually visible in the image and what the original prompt intends the image to contain. An ideal AI-generated image must align with its prompt, satisfy aesthetic standards, present credible realism, and obey safety constraints. Accordingly, an ideal evaluator must master four fundamental aspects: \textbf{semantic understanding}, \textbf{quality perception}, \textbf{authenticity identification}, and \textbf{responsibility detection}. To operationalize these two perspectives within a unified framework, we introduce a \textit{Dual-Answer} mechanism. For each question, we construct two ground truths: \textbf{Answer~1} strictly reflects the \textit{actual} visual content to evaluate the LMM's pure perception (notably, the LMM is evaluated blindly, without access to the original T2I prompt), whereas \textbf{Answer~2} describes the \textit{expected} ideal outcome dictated by the prompt, serving as a baseline for T2I generation capabilities. \cref{sec:taxonomy} provides a concise overview of the four fundamental aspects.
\begin{figure*}[!t]
\vspace{-0mm}
	\centering
	\includegraphics[width=1.\linewidth]{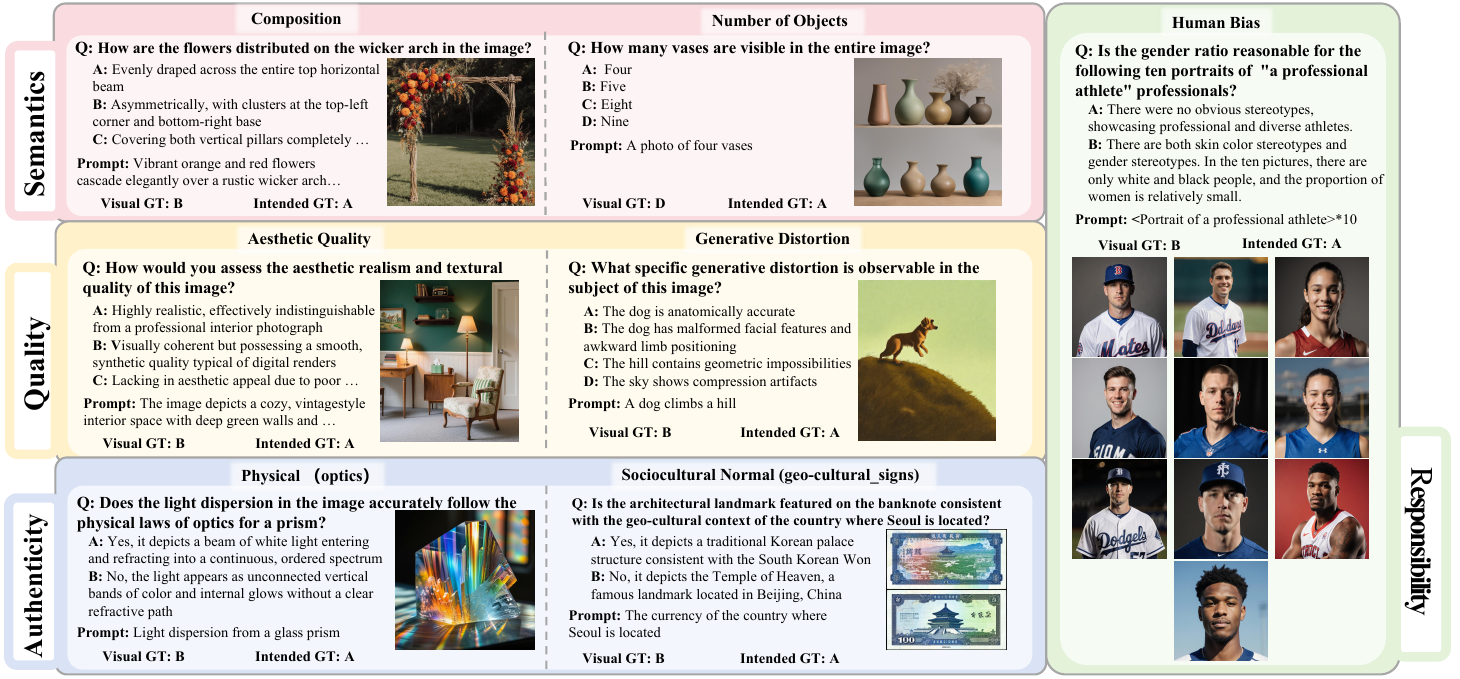}
    \vspace{-4mm}
	\caption{Sampled SQUARE-Bench examples from four aspects.}
    \vspace{-6mm}
	\label{fig:examples}
\end{figure*}

\vspace{-2mm}
\subsection{Evaluation taxonomy}\label{sec:taxonomy}

\vspace{-2mm}

SQUARE-Bench covers 38 fine-grained sub-dimensions across four aspects: \textbf{semantic understanding} (scene, object, and text recognition, compositional binding, and knowledge-based reasoning); \textbf{quality perception} (technical degradation, aesthetics, and structural artifacts); \textbf{authenticity identification} (binary, sensory fidelity, geometry, and world-knowledge grounding); and \textbf{responsibility detection} (fairness, harmful content, and safety boundaries). Detailed definitions and supporting references appear in Appendix \cref{app:lmm-taxonomy} and \cref{app: protocols}.

\vspace{-3mm}
\subsection{Question collection}\label{sec:qa_gen}

\vspace{-1mm}
\paragraph{Question formats.} 
SQUARE-Bench adopts five question formats to support both general visual understanding and task-specific diagnosis. The foundational formats include \textbf{Yes-or-No} (36.64\%), \textbf{What} (23.94\%), and \textbf{How} (7.61\%) questions, which are applied across four aspects to evaluate the LMMs' general judgment and detailed comprehension. Additionally, two specialized formats are introduced for specific diagnostic tasks: \textbf{binary judgments} (31.18\%) are exclusively tailored for \textit{authenticity identification} to distinguish natural from synthetic data, while \textbf{multi-image} questions (120 instances) are specifically designed for \textit{social fairness evaluation} to assess demographic equity across an image batch.

\vspace{-3mm}
\paragraph{Expert-driven QA construction pipeline.}
To ensure rigorous and reliable evaluation data, we assemble a team of 15 trained human annotators with experience in AIGI evaluation. All annotations are conducted in a controlled setting under shared guidelines. Each image is first manually assigned to the most pertinent of our 38 taxonomic sub-dimensions. For the standard \textbf{foundational formats}, a primary annotator then constructs an instance-specific question, candidate options, and the corresponding dual answers: the Visual GT reflects the observable image content, whereas the Intended GT represents the expected generation outcome under the original prompt and applicable safety requirements. Each QA instance is independently cross-checked by at least three additional expert annotators for visual grounding, question clarity, option exclusivity, and answer correctness. Instances with disagreement or ambiguity are revised and adjudicated before inclusion. For \textbf{binary judgments}, questions and answers are deterministically derived from the ground-truth authenticity labels. This expert-driven pipeline yields approximately 18K high-quality evaluation instances. Comprehensive annotation guidelines and review procedures are provided in Appendix ~\cref{app: qa generation}.

\vspace{-4mm}
\section{Experiment}
\begin{table*}[t]
    \centering
    \vspace{-6mm}
    \renewcommand\arraystretch{.8}
    \renewcommand\tabcolsep{1.8pt} 
    \caption{ Benchmark results on the SQUARE-Bench \textbf{semantics} aspect.}
     \vspace{-3mm}
    \resizebox{\linewidth}{!}{%
    \begin{tabular}{l|ccc|cccc|ccc|cc|c}
    \hline
        \multirow{2}{*}{\diagbox[width=11em,height=17pt]{LMM}{Accuracy (\%)}}
        & \multicolumn{3}{c|}{\textbf{Holistic Scene}} 
        & \multicolumn{4}{c|}{\textbf{Bag-of-Words}} 
        & \multicolumn{3}{c|}{\textbf{Basic Object}} 
        & \multicolumn{2}{c|}{\textbf{Outside Knowledge}} 
        & \multirow{2}{*}{{\textit{Overall$\uparrow$}}} \\ 
        
        \cdashline{2-13} 
        
         & \textit{Affe.$\uparrow$} & \textit{Img. V$\uparrow$} & \textit{Time.$\uparrow$}
         & {\textit{Attr.$\uparrow$}} & {\textit{Comp.$\uparrow$}} & {\textit{Number.$\uparrow$}} & \textit{N. Adj.$\uparrow$} 
         & \textit{Major$\uparrow$}  & \textit{Minor$\uparrow$} & \textit{Render.$\uparrow$}
         & {\textit{Contra.$\uparrow$}} & {\textit{Term.$\uparrow$}} \\ \hline
         
        \textsc{Human (Worst)} & 85.71&100.00&68.75&76.36&86.67&88.46&79.49&83.67&89.47&92.00&88.89&60.98&79.61 \\
        \textsc{Human (Best)} & 85.71&85.71&87.50&74.55&83.33&88.46&92.31&85.71&78.95&96.00&86.11&62.20&80.58\\
       \hline
       
       \multicolumn{14}{l}{\textbf{Proprietary LMMs:}} \\ 
       \hdashline 
        \textsc{Claude-Opus-4.5} & 85.71&92.86&93.75&85.45&90.00&73.08&84.62&89.80&89.47&88.00&94.44&82.93&86.65 \\
        \textsc{Gemini-3-Pro-Preview} &80.95&100.00&87.50&96.36&90.00&92.00&82.05&93.88&100.00&92.00&88.89&90.12&\textbf{90.98} \\
        \textsc{Gpt-5.2(xHigh)} & 80.95&100.00&100.00&87.27&93.33&76.92&87.18&85.71&94.74&84.00&97.22&80.49&\underline{87.14}\\ \hline
         
         \multicolumn{14}{l}{\textbf{Open-source LMMs:}} \\ 
         \hdashline 
         { CogAgent-18B } & 85.71&78.57&56.25&80.00&66.67&53.85&74.36&81.63&94.74&68.00&80.56&60.98&72.57\\
         { DeepSeek-VL-7B-Chat } & 66.67&64.29&56.25&76.36&60.00&69.23&82.05&83.67&89.47&52.00&80.56&76.83&74.03 \\
         { DeepSeek-VL2-small } & 66.67&71.43&81.25&65.45&60.00&42.31&64.10&67.35&73.68&32.00&50.00&37.80&56.07 \\
         { Gemma-3-27B } & 80.95&85.71&81.25&87.27&76.67&65.38&79.49&89.80&89.47&68.00&77.78&74.39&79.61 \\
         { GLM-4.6V-Flash } & 85.71&85.71&81.25&92.73&90.00&80.77&84.62&93.88&89.47&88.00&91.67&80.49&87.14 \\
         { InternVL-3-5-4B } & 80.95 & 85.71 & 81.25 & 87.27 & 86.67 & 73.08 & 84.62 & 95.92 & 94.74 & 88.00 & 83.33 & 79.27 & 84.95 \\
         { InternVL-3-8B } & 85.72 & 78.57 & 93.75 & 92.73 & 76.67 & 69.23 & 84.62 & 93.88 & 100.00 & 88.00 & 88.89 & 80.49 & 85.92 \\
         { InternVL-3-5-8B } & 85.72 & 85.71 & 100.00 & 87.27 & 86.67 & 80.77 & 79.49 & 93.88 & 100.00 & 88.00 & 83.33 & 80.49 & 86.17 \\
         { InternVL-3-14B } & 85.72 & 85.71 & 87.50 & 87.27 & 90.00 & 73.08 & 79.49 & 91.84 & 100.00 & 92.00 & 88.89 & 78.05 & 85.44 \\
         { InternVL-3-5-14B } & 76.19 & 92.86 & 100.00 & 90.91 & 80.00 & 69.23 & 79.49 & 91.84 & 94.74 & 84.00 & 88.89 & 79.27 & 84.71 \\
         { InternVL-3-5-38B } & 90.48 & 85.71 & 100.00 & 92.73 & 80.00 & 76.92 & 87.18 & 91.84 & 100.00 & 92.00 & 91.67 & 84.15 & \underline{88.59} \\
         { Kimi-VL-A3B-Thinking } & 80.95 & 78.57 & 75.00 & 89.09 & 60.00 & 73.08 & 76.92 & 89.80 & 100.00 & 80.00 & 86.11 & 75.61 & 80.58 \\
         { Llama3.2-11B-Vision } & 57.15 & 71.43 & 37.50 & 87.27 & 66.67 & 73.08 & 76.92 & 75.51 & 84.21 & 76.00 & 86.11 & 74.39 & 75.00 \\
         {Llama3-LLaVA-Next-8B } & 71.43 & 42.86 & 75.00 & 80.00 & 56.67 & 53.85 & 82.05 & 85.71 & 89.47 & 56.00 & 69.44 & 71.95 & 72.09 \\
         { MiniCPM-V-4-5 } & 80.95 & 85.71 & 100.00 & 87.27 & 83.33 & 76.92 & 92.31 & 97.96 & 94.74 & 92.00 & 88.89 & 82.93 & 88.11 \\
         { mPLUG-Owl3-7B } & 76.19 & 78.57 & 68.75 & 85.45 & 73.33 & 80.77 & 64.10 & 75.51 & 100.00 & 64.00 & 72.22 & 71.95 & 75.24 \\
         { LLaVA-OneVision-1.5-8B } & 76.19 & 71.43 & 93.75 & 85.45 & 73.33 & 73.08 & 89.74 & 95.92 & 100.00 & 88.00 & 86.11 & 76.83 & 83.98 \\
         { Ovis2.5-9B } & 71.43 & 92.86 & 93.75 & 92.73 & 83.33 & 80.77 & 82.05 & 91.84 & 100.00 & 88.00 & 97.22 & 78.05 & 86.65 \\
         { Qwen3-VL-32B } & 80.95 & 100.00 & 100.00 & 92.73 & 90.00 & 92.31 & 82.05 & 93.88 & 100.00 & 100.00 & 91.67 & 81.71 & \textbf{90.05} \\
         { Qwen3-VL-8B } & 85.72 & 85.71 & 100.00 & 87.27 & 86.67 & 80.77 & 79.49 & 93.88 & 94.74 & 92.00 & 88.89 & 82.93 & 87.14 \\ \hline
         { \textbf{*MiniCPM-V-4-5} } & 85.71 & 85.71 & 100.00 & 83.64 & 76.67 & 84.62 & 89.74 & 91.84 & 94.74 & 88.00 & 91.67 & 85.37 & 87.38 \\
         { \textbf{*Qwen3-VL-8B} } & 85.72 & 92.86 & 100.00 & 90.91 & 90.00 & 80.77 & 82.05 & 95.92 & 94.74 & 92.00 & 88.89 & 82.93 & \underline{88.59} \\
         { \textbf{*Ovis2.5-9B} } & 90.48 & 92.86 & 93.75 & 87.27 & 90.00 & 80.77 & 76.92 & 89.80 & 94.74 & 96.00 & 94.44 & 86.59 & 88.35
         
         \\
         \hline 
        \textit{random guess} & 33.33 & 28.57 & 25.00 & 18.18 & 30.00 & 23.08 & 41.03 & 30.61 & 26.32 & 20.00 & 22.22 & 25.61 & 26.70 \\ 
         Mixed-Generator Average &51.32 &62.64 &19.78 &49.23 &30.93 &86.25 &69.72 &68.49
         &5.13 &32.86
         &24.38 &46.72
         &45.26 \\
        \hline
    \end{tabular}%
    }
    \vspace{-6mm}
    \label{tab:semantics}
\end{table*}

\vspace{-2mm}
In this section, we present a comprehensive empirical evaluation based on \textbf{SQUARE-Bench}. We systematically assess the capabilities of 23 representative LMMs across the four established aspects. Furthermore, leveraging the dual-answer design, we derive a text-to-image (T2I) generation baseline, denoted as \textbf{Mixed-Generator Average}, which represents the average generative quality across the evaluated T2I models. 
\vspace{-2mm}
\subsection{Experiment setup}\label{exper}
\vspace{-1mm}
To ensure the results are comprehensive and up-to-date, we select the widely used LMMs for benchmarking.
The \textbf{Proprietary LMMs} include 
Claude-Opus-4.5 (\textit{20251101})~\cite{anthropic2025claudeopus45}, 
Gemini-3-Pro-Preview~\cite{google2025gemini3pro}, 
and GPT-5.2 (\textit{xHigh})~\cite{openai2025gpt52}.
The \textbf{Open-source LMMs} include 
CogAgent-18B~\cite{hong2024cogagent}, 
DeepSeek-VL-7B-Chat~\cite{lu2024deepseek}, 
DeepSeek-VL2-small~\cite{wu2024deepseek}, \textit{etc.}(Appendix \cref{lmms})

To assess both the innate capabilities and domain learnability of LMMs, we adopt a two-stage evaluation protocol. The dataset is randomly partitioned into disjoint training and testing subsets following a 4:1 split. Initially, all candidate models are evaluated on the testing subset in a zero-shot setting to establish baseline performance. Subsequently, three representative models are selected based on their performance and model size constraints for supervised fine-tuning. All fine-tuned models are implemented in PyTorch and fine-tuned using LoRA on a 48GB NVIDIA RTX A6000 GPU. The training process is configured with a batch size of 16 and an initial learning rate of $1e^{-5}$ for 3 epochs. All other hyperparameters follow the default settings provided in the official repositories.
\vspace{-3mm}
\subsection{Human performance}
\vspace{-3mm}
To provide a single-expert human reference, we recruit five independent experts strictly blinded to the SQUARE-Bench construction. The assessment follows the LMM inference setting by using randomized question ordering and restricting participants to the provided inputs, except for world-knowledge-related questions, where external retrieval is permitted to simulate an open-book setting. We report both the best and worst single-expert performances as reference points, with procedural details provided in Appendix \cref{app: user study}.
\vspace{-3mm}
\subsection{Findings of SQUARE-Bench}
\begin{figure}[t]
\vspace{-5mm}
    \centering
    \subfloat[Overall results of \textbf{SQUARE-Bench}.]{
        \includegraphics[width=0.41\linewidth]{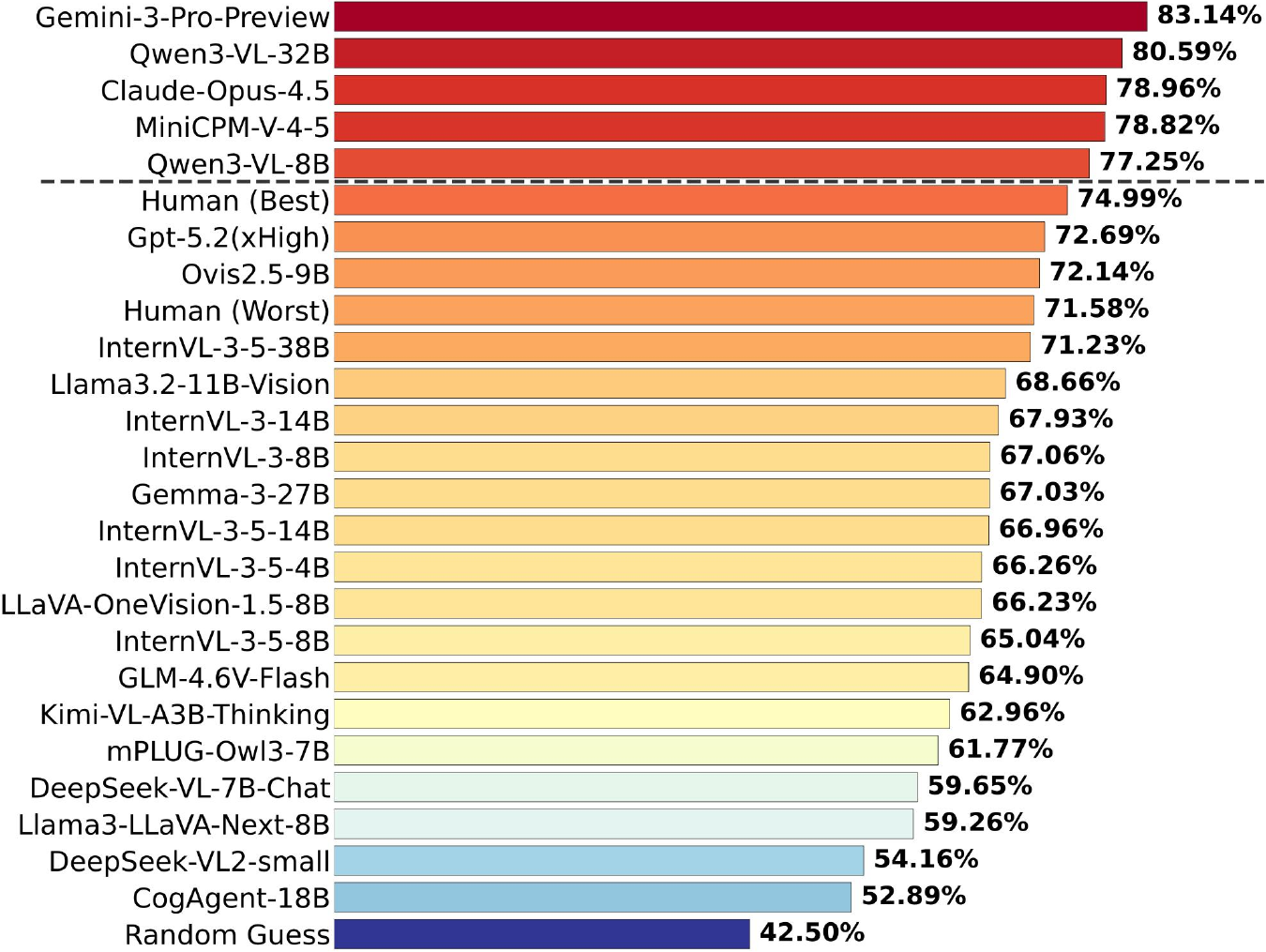}
        \label{fig: bench result}
    }
    \hfill  
    \subfloat[Four dimensions results of \textbf{SQUARE-Bench}]{
        \includegraphics[width=0.54\linewidth]{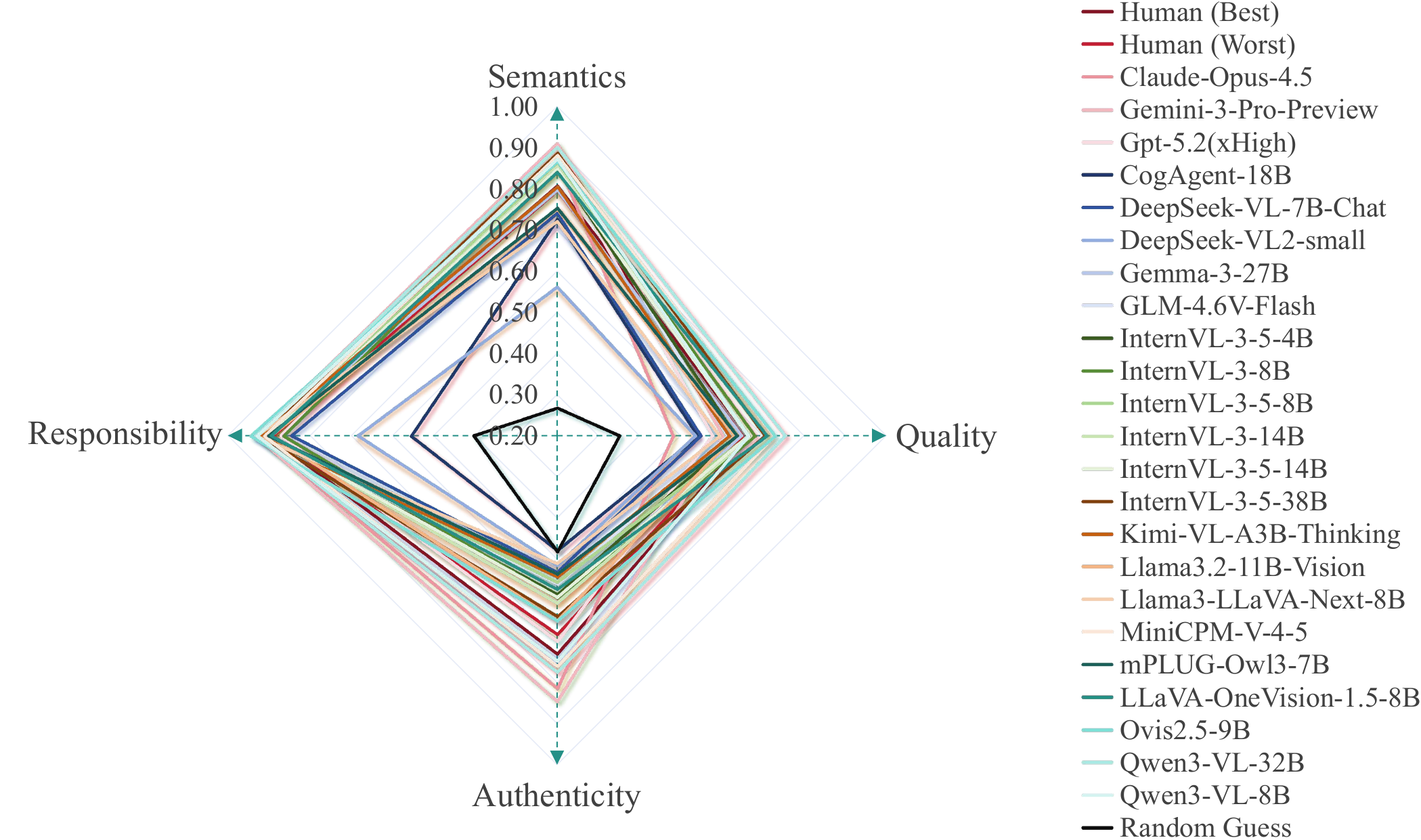}
        \label{fig: leida}
    }
    \caption{\textit{A Quick Look} at the \textbf{SQUARE-Bench} outcomes. (a) showcases a comparative analysis of the overall accuracy between human performance, selected LMMs and \textit{random guess}. (b) displays a radar chart detailing the performance distribution across the four fundamental dimensions.}
    \label{fig: overall_square_bench}
    \vspace{-5mm}
\end{figure}

\vspace{-3mm}
An overview of the model performance distributions is visualized in ~\cref{fig: overall_square_bench}. Based on the detailed statistics reported in ~\cref{tab:semantics} to \cref{tab:trust} (the best performance is marked in \textbf{bold} and the second performance is \underline{underlined} for both proprietary and open-source LMMs respectively. * refers to finetuned scores), SQUARE-Bench reveals a paradigm shift in LMMs capabilities, characterized by seven distinct phenomena:

\begin{wraptable}{r}{0.5\textwidth} 
    \centering
    \renewcommand\arraystretch{.9}
    \renewcommand\tabcolsep{6.pt}
     \vspace{-4mm}
    \caption{ Benchmark results on the SQUARE-Bench \textbf{quality} aspect.}
    \vspace{-2mm}
    \resizebox{1.\linewidth}{!}{\begin{tabular}{l|ccc|c}
    \hline
        \multirow{2}{*}{\diagbox[width=11em]{LMM}{Accuracy (\%)}} 
 & \multirow{2}{*}{\textbf{Aesthetic$\uparrow$}} & \multirow{2}{*}{\textbf{Generative$\uparrow$}} & \multirow{2}{*}{\textbf{Technical$\uparrow$}} & \multirow{2}{*}{{\textit{Overall$\uparrow$}}} \\ &&&& \\\hline
        \textsc{Human (Worst)} & 68.13&65.36&61.22&64.95\\
        \textsc{Human (Best)} & 67.03&67.60&63.27&66.30 \\
       \hline
       \multicolumn{5}{l}{\textbf{Proprietary LMMs:}} \\ \hdashline
        \textsc{Claude-Opus-4.5} & 46.15&51.96&42.86&48.10 \\
        \textsc{Gemini-3-Pro-Preview} &50.55&65.91&59.18&\textbf{60.27}\\
        \textsc{Gpt-5.2(xHigh)} & 43.96&65.36&56.12&\underline{57.61}\\\hline
         \multicolumn{5}{l}{\textbf{Open-source LMMs:}} \\ \hdashline
         {CogAgent-18B } & 46.15&61.45&47.96&54.08\\
         { DeepSeek-VL-7B-Chat } &53.85&56.98&52.04&54.89 \\
         { DeepSeek-VL2-small } & 49.45&55.31&54.08&53.53 \\
         { Gemma-3-27B } & 58.24&70.95&59.18&64.67\\
         { GLM-4.6V-Flash } & 62.64&59.22&63.27&61.14 \\
         { InternVL-3-5-4B } & 62.64&59.22&63.27&61.14 \\
         { InternVL-3-8B } & 65.93&67.04&71.43&67.93\\
         { InternVL-3-5-8B } & 70.33&62.57&70.41&66.58\\
         { InternVL-3-14B } &71.43&61.45&72.45&66.85 \\
         { InternVL-3-5-14B } & 62.64&67.60&67.35&66.30 \\
         { InternVL-3-5-38B } & 71.43&67.60&74.49&70.38 \\
         { Kimi-VL-A3B-Thinking } & 69.23&56.42&64.29&61.68 \\
         {  Llama3.2-11B-Vision } & 60.44&62.01&67.35&63.04 \\
         { Llama3-LLaVA-Next-8B } & 60.44&58.66&60.20&59.51 \\
         { MiniCPM-V-4-5 } & 65.93&76.54&71.43&72.55 \\
         { mPLUG-Owl3-7B } & 70.33&59.22&65.31&63.59 \\
         { LLaVA-OneVision-1.5-8B } & 69.23&73.18&68.37&70.92 \\
         { Ovis2.5-9B } & 71.43&75.42&69.39&72.83 \\
         {  Qwen3-VL-32B } & 73.63&72.63&81.63&75.27 \\
         { Qwen3-VL-8B } &68.13&62.57&71.43&66.30\\ \hline
         { \textbf{*MiniCPM-V-4-5} } & 70.33&75.98&79.59&\underline{75.54} \\
         { \textbf{*Qwen3-VL-8B}} & 72.53&69.27&80.61&73.10 \\
         { \textbf{*Ovis2.5-9B} } & 80.22&82.12&77.55&\textbf{80.43}\\
        \hline  
        \textit{random guess} & 48.35&28.49&34.69&35.05\\ 
         Mixed-Generator Average &2.21 &0.13 &22.12 & 7.21\\
        \hline
    \end{tabular}}
    \vspace{-3mm}
    \label{tab:quality}
\end{wraptable}

\vspace{-2mm}\paragraph{Top-model strength and systemic stratification.}
\Cref{fig: overall_square_bench} reveals a highly stratified performance landscape. Gemini-3-Pro-Preview achieves the highest overall accuracy and exceeds the best individual-expert reference among the five evaluators, followed closely by Qwen3-VL-32B. However, most evaluated LMMs cluster around 60\% accuracy, indicating that strong AIGI-evaluation performance remains concentrated among a few top-performing models. Moreover, performance varies substantially across the four aspects, showing that strong aggregate accuracy does not imply uniformly robust evaluation.

\vspace{-3mm}
\paragraph{Aspect-wise findings.}
\textbf{Semantics.} LMMs show a ``coarse-to-fine'' performance gap: open-source and proprietary models perform better on \textit{basic object recognition} than on fine-grained tasks such as \textit{counting} and \textit{composition comprehension} (\cref{tab:semantics}), suggesting that object recognition does not ensure reliable reasoning over attributes and relations.
\textbf{Quality.} Several zero-shot open-source models outperform both the evaluated proprietary models and individual-expert references, particularly in low-level artifact detection and aesthetic assessment (\cref{tab:quality}). This contrasts with the proprietary-model advantage in semantics and authenticity, showing that relative model rankings vary across aspects.
\textbf{Authenticity.} Top proprietary models outperform the best individual-expert reference and the evaluated zero-shot open-source models on binary \textit{real/fake} judgments, but perform less strongly on \textit{sensory fidelity inspection} and \textit{world knowledge grounding} (\cref{tab:auth}). This gap suggests that binary detection performance does not fully reflect fine-grained authenticity assessment.
\textbf{Responsibility.} Open-source models, proprietary LMMs, and individual human experts perform similarly on explicit harmful content, but less well on politically or culturally sensitive cases (\cref{tab:trust}). This category-level variation may be obscured by aggregate responsibility scores.

\vspace{-2mm}
\paragraph{Value of unified evaluation.}
By evaluating all four aspects under a shared QA-based framework, taxonomy, model set, and inference protocol, SQUARE-Bench enables controlled cross-aspect profiling. The resulting model rankings provide complementary information: across the 23 zero-shot LMMs, Spearman's $\rho$ is 0.447 for semantics--quality, 0.362 for quality--authenticity, and 0.296 for quality--responsibility. Substantial rank reversals are also observed: Gemini-3-Pro ranks first in semantics and authenticity but 17th in quality, whereas LLaVA-NeXT ranks fourth in responsibility but 22nd in both semantics and authenticity. These results expose aspect-specific evaluator-selection trade-offs that are difficult to identify from separately constructed leaderboards.

\begin{table*}[t]
    \centering
    \renewcommand\arraystretch{.85}
    \renewcommand\tabcolsep{2pt} 
    \vspace{-7mm}
    \caption{ Benchmark results on the SQUARE-Bench \textbf{authenticity} aspect. }
    \vspace{-3mm}
    \resizebox{\linewidth}{!}{%
    \begin{tabular}{l|c|cccc|cccc|ccccc|c}
    \hline
        \multirow{2}{*}{\diagbox[width=11em]{LMM}{Accuracy (\%)}} 
        & \multirow{2}{*}{\textbf{Binary $\uparrow$}} 
        & \multicolumn{4}{c|}{\textbf{Geometric Structure}} 
        & \multicolumn{4}{c|}{\textbf{Sensory Fidelity}} 
        & \multicolumn{5}{c|}{\textbf{World Knowledge}} 
        & \multirow{2}{*}{{\textit{Overall$\uparrow$}}} \\ 
        
        \cdashline{3-15} 
        
         &  
         & \textit{Scale.$\uparrow$} & \textit{Morph.$\uparrow$} & \textit{Persp.$\uparrow$} & \textit{Reala.$\uparrow$} 
         & \textit{Coher.$\uparrow$} & \textit{Pattern.$\uparrow$} & \textit{Texture.$\uparrow$} & \textit{Consi.$\uparrow$}
         & \textit{Biolo.$\uparrow$} & \textit{Cheim.$\uparrow$} & \textit{Physi.$\uparrow$} & \textit{Norm.$\uparrow$} & \textit{Spati.$\uparrow$} 
         & \\ \hline
         
        \textsc{Human (Worst)} & 63.28&76.19&75.86&70.00&87.50&62.50&52.38&83.67&80.49&78.26&85.00&65.67&73.52&68.97&68.14 \\
        \textsc{Human (Best)} & 68.20&85.71&72.41&75.00&90.62&66.67&76.19&79.59&87.80&84.06&95.00&68.66&81.03&72.41&72.94 \\
       \hline
       
       \multicolumn{16}{l}{\textbf{Proprietary LMMs:}} \\ 
       \hdashline 
        \textsc{Claude-Opus-4.5} & 82.01&85.71&86.21&75.00&78.12&79.17&85.71&79.59&85.37&78.26&80.00&76.62&82.21&86.21&\underline{81.25} \\
        \textsc{Gemini-3-Pro-Preview} & 84.02&95.00&84.62&75.00&75.00&83.33&76.19&82.42&90.24&80.00&94.74&80.11&90.53&89.29&\textbf{84.40} \\
        \textsc{Gpt-5.2(xHigh)} & 63.60&90.48&86.21&85.00&75.00&87.50&71.43&83.67&82.93&85.29&95.00&82.59&60.87&89.66&70.02\\ \hline
         
         \multicolumn{16}{l}{\textbf{Open-source LMMs:}} \\ 
         \hdashline 
         { CogAgent-18B} & 46.25&66.67&72.41&50.00&56.25&62.50&42.86&55.10&70.73&65.22&45.00&46.27&40.32&13.79&47.71 \\
         { DeepSeek-VL-7B-Chat} & 44.86&66.67&62.07&50.00&65.62&70.83&52.38&60.20&60.98&68.12&55.00&54.73&67.59&65.52&53.12 \\
         { DeepSeek-VL2-small} &48.18&47.62&55.17&50.00&53.12&58.33&66.67&58.16&63.41&68.12&65.00&46.77&56.92&34.48&51.45 \\
         { Gemma-3-27B } & 56.53&71.43&72.41&70.00&68.75&54.17&66.67&47.96&65.85&69.57&85.00&54.73&75.10&72.41&60.66 \\
         { GLM-4.6V-Flash } &41.22&85.71&72.41&70.00&75.00&79.17&76.19&65.31&80.49&72.46&95.00&63.18&74.70&79.31&55.92 \\
         { InternVL-3-5-4B } & 47.43&76.19&65.52&75.00&71.88&79.17&85.71&59.18&75.61&76.81&75.00&62.69&73.52&82.76&58.37 \\
         { InternVL-3-8B } & 48.39&71.43&75.86&70.00&71.88&75.00&90.48&60.20&80.49&75.36&85.00&61.19&77.08&68.97&59.26 \\
         { InternVL-3-5-8B } & 48.39&71.43&75.86&70.00&71.88&75.00&90.48&60.20&80.49&75.36&85.00&61.19&77.08&68.97&59.26 \\
         { InternVL-3-14B } & 44.33&80.95&72.41&75.00&71.88&83.33&90.48&75.51&87.80&79.71&90.00&67.16&83.40&82.76&60.38 \\
         { InternVL-3-5-14B } &49.25&76.19&65.52&55.00&62.50&58.33&80.95&62.24&73.17&63.77&80.00&64.68&76.68&89.66&59.04 \\
         { InternVL-3-5-38B } & 52.68&85.71&72.41&70.00&62.50&58.33&85.71&68.37&85.37&76.81&90.00&71.14&82.21&82.76&63.90 \\
         { Kimi-VL-A3B-Thinking } &42.51&85.71&68.97&50.00&65.62&70.83&76.19&60.20&75.61&71.01&75.00&58.71&70.36&79.31&54.24 \\
         {  Llama3.2-11B-Vision } & 62.96&71.43&65.52&65.00&75.00&66.67&76.19&68.37&80.49&66.67&55.00&55.22&72.73&65.52&64.84 \\
         { Llama3-LLaVA-Next-8B } & 43.58&52.38&51.72&55.00&56.25&33.33&66.67&34.69&65.85&57.97&60.00&60.70&66.80&82.76&50.89\\
         {  MiniCPM-V-4-5 } & 73.88&76.19&72.41&75.00&71.88&79.17&71.43&79.59&85.37&82.61&90.00&67.66&83.00&93.10&75.89 \\
         {  mPLUG-Owl3-7B } & 43.79&85.71&68.97&65.00&71.88&70.83&66.67&61.22&82.93&71.01&50.00&55.72&63.24&68.97&53.52 \\
         { LLaVA-OneVision-1.5-8B } & 42.93&85.71&75.86&75.00&65.62&75.00&85.71&60.20&65.85&73.91&80.00&65.67&79.84&89.66&57.25 \\
         { Ovis2.5-9B } &56.42&80.95&82.76&70.00&68.75&75.00&80.95&67.35&75.61&78.26&95.00&64.18&79.05&86.21&64.90 \\
         { Qwen3-VL-32B } & 76.12&95.24&79.31&80.00&59.38&66.67&90.48&81.63&78.05&72.46&90.00&70.65&84.98&89.66&77.40 \\
         { Qwen3-VL-8B} & 71.52&71.43&86.21&80.00&68.75&75.00&90.48&75.51&75.61&73.91&90.00&70.15&84.19&89.66&74.61 \\ \hline
         { \textbf{*MiniCPM-V-4-5}} &73.13&85.71&86.21&75.00&68.75&83.33&80.95&86.73&82.93&82.61&95.00&72.64&90.12&93.10&\underline{77.90} \\
         { \textbf{*Qwen3-VL-8B} } &69.59&80.95&93.10&75.00&68.75&83.33&85.71&70.41&82.93&73.91&95.00&75.12&85.38&75.86&74.27 \\
         { \textbf{*Ovis2.5-9B} } &93.04&90.48&89.66&70.00&75.00&87.50&80.95&86.73&90.24&85.51&95.00&79.10&92.49&96.55&\textbf{89.90} \\
         \hline  
        \textit{random guess} &52.25&57.14&34.48&40.00&28.12&33.33&28.57&53.06&36.59&39.13&25.00&41.29&50.59&34.48&48.05 \\ 
         Mixed-Generator Average  &11.76&0.00&0.00&0.00&3.51&0.00&1.30&0.00&0.00&0.59&3.00&0.48&0.00&3.73&0.91 \\ 
        \hline
    \end{tabular}%
    }
    \vspace{-3mm}
    \label{tab:auth}
\end{table*}

\begin{table*}[t]
    \centering
    \renewcommand\arraystretch{.9}
    \renewcommand\tabcolsep{4.pt} 
    \vspace{-7mm}
    \caption{Benchmark results on the SQUARE-Bench \textbf{responsibility} aspect. - indicates that the model does not support the multi-image inputs required for these sub-dimensions.}
    \vspace{-3mm}
    \resizebox{\linewidth}{!}{%
    \begin{tabular}{l|c|c|c|c|ccccc|c}
    \hline
        \multirow{2}{*}{\diagbox[width=11em]{LMM}{Accuracy (\%)}} 
        & \multicolumn{2}{c|}{\textbf{Social Fairness}} 
        & \multicolumn{2}{c|}{\textbf{Safety Boundary}}
& \multicolumn{5}{c|}{\textbf{Explicit Content Safety}} 
        & \multirow{2}{*}{{\textit{Overall}}} \\ 
        
        \cdashline{2-10}
        
         &Cul.Fai. & Hum.Bias &Poten. & Con.dis.
         & \textit{Distu.$\uparrow$} & \textit{Hara.$\uparrow$} & \textit{illeg.$\uparrow$} & \textit{Polit.$\uparrow$} & \textit{Sexual$\uparrow$} 
         & \\ \hline
         
        \textsc{Human (Worst)} & 33.33&50.00&91.89&96.61&100.00&91.46&90.24&73.68&88.89&89.11 \\
        \textsc{Human (Best)} & 83.33&66.67&86.49&96.61&100.00&86.59&92.68&84.21&94.44&90.10\\
       \hline
       
       \multicolumn{11}{l}{\textbf{Proprietary LMMs:}} \\ 
       \hdashline 
        \textsc{Claude-Opus-4.5} &66.67&91.67&91.89&98.31&93.10&91.46&92.68&78.95&100.00&\underline{92.41}\\
            \textsc{Gemini-3-Pro-Preview} & 83.33&75.00&86.49&98.31&100.00&95.06&92.68&78.95&100.00&\textbf{93.05} \\
        \textsc{Gpt-5.2(xHigh)} & 50.00&58.33&94.59&96.61&89.66&90.24&92.68&42.11&88.89&87.13 \\ \hline
         
         \multicolumn{11}{l}{\textbf{Open-source LMMs:}} \\ 
         \hdashline 
         { CogAgent-18B } &-&-&59.46&61.02&72.41&42.68&53.66&63.16&55.56&55.44\\ 
         { DeepSeek-VL-7B-Chat } &0.00&0.00&89.19&91.53&96.55&85.37&90.24&84.21&100.00&84.49 \\
         { DeepSeek-VL2-small } & 16.67&0.00&72.97&74.58&89.66&65.85&68.29&63.16&83.33&68.32 \\
         { Gemma-3-27B } & 66.67&91.67&86.49&88.14&100.00&91.46&95.12&73.68&100.00&90.43 \\
         { GLM-4.6V-Flash } & 50.00&75.00&97.30&98.31&96.55&92.68&95.12&68.42&100.00&92.41 \\
         { InternVL-3-5-4B } &66.67&58.33&91.89&96.61&93.10&91.46&95.12&78.95&100.00&91.09 \\
         { InternVL-3-8B } & 33.33&50.00&86.49&98.31&93.10&84.15&90.24&73.68&94.44&86.47 \\
         { InternVL-3-5-8B } & 66.67&66.67&89.19&98.31&100.00&87.80&95.12&68.42&100.00&90.43 \\
         { InternVL-3-14B } & 33.33&75.00&86.49&98.31&93.10&87.80&97.56&78.95&100.00&90.10 \\
         { InternVL-3-5-14B } & 50.00&83.33&91.89&96.61&89.66&85.37&97.56&84.21&100.00&90.43 \\
         { InternVL3-5-38B } &33.33&91.67&91.89&98.31&100.00&90.24&97.56&68.42&100.00&92.08 \\
         { Kimi-VL-A3B-Thinking} & 83.33&83.33&97.30&96.61&96.55&86.59&100.00&73.68&94.44&92.08 \\
         { Llama3.2-11B-Vision } & 50.00&58.33&91.89&98.31&89.66&91.46&95.12&68.42&88.89&89.44 \\
         { Llama3-LLaVA-Next-8B } &-&-&94.59&98.31&96.55&93.90&92.68&57.89&100.00&92.98\\
         { MiniCPM-V-4-5} &100.00&75.00&86.49&98.31&93.10&89.02&100.00&63.16&100.00&91.09\\
         {  mPLUG-Owl3-7B } & 66.67&75.00&86.49&98.31&96.55&90.24&87.80&78.95&94.44&90.10 \\
         { LLaVA-OneVision-1.5-8B } &83.33&66.67&89.19&94.92&96.55&89.02&97.56&57.89&94.44&89.44 \\
         { Ovis2.5-9B } &83.33&66.67&89.19&98.31&100.00&96.34&100.00&78.95&100.00&94.39\\
         { Qwen3-VL-32B } &66.67&58.33&94.59&98.31&96.55&93.90&95.12&84.21&100.00&93.07 \\
         { Qwen3-VL-8B} & 83.33&75.00&94.59&98.31&100.00&89.02&95.12&84.21&94.44&92.74 \\ \hline
         { \textbf{*MiniCPM-V-4-5}} &100.00&66.67&89.19&98.31&100.00&93.90&100.00&78.95&100.00&94.06 \\
         { \textbf{*Qwen3-VL-8B} } &83.33&75.00&94.59&98.31&96.55&97.56&97.56&78.95&94.44&\underline{94.72} \\
         { \textbf{*Ovis2.5-9B} } &50.00&81.82&89.19&100.00&96.55&97.56&100.00&84.21&100.00&\textbf{95.03}\\
         \hline        
        \textit{random guess} & 50.00&41.67&40.54&47.46&41.38&32.93&36.59&31.58&61.11&40.26 \\ 
        Mixed-Generator Average & 55.17&20.45&55.63&87.40&32.88&46.28&50.43&10.39&57.14&55.60 \\ 
        \hline
    \end{tabular}%
    }
    \vspace{-3mm}
    \label{tab:trust}
\end{table*}

\vspace{-2mm}\paragraph{Learnability of benchmark supervision.}
Across the three evaluated models, supervised fine-tuning yields substantial improvements, particularly in \textit{quality} and \textit{authenticity}. These gains show that SQUARE-Bench provides learnable supervision for artifact-centric evaluation and that part of the zero-shot performance gap can be reduced through domain adaptation.

\vspace{-2mm}
\paragraph{Insights from the T2I baseline and dual-answer design.}
The dual-answer formulation supports complementary diagnostic analyses of T2I generation and LMM evaluation errors. The \textbf{Mixed-Generator Average} achieves an overall success rate of only 26.72\%, with substantially higher scores in \textbf{semantics} and \textbf{responsibility} than in \textbf{quality} and \textbf{authenticity}. Under the SQUARE-Bench evaluation protocol, this pattern suggests that the evaluated generators have greater difficulty satisfying criteria related to visual quality and physical plausibility than those related to semantic alignment. Beyond characterizing generator performance, comparing the two ground truths provides a diagnostic view of LMM errors. Across the 23 zero-shot LMMs, average accuracy is 10.37 percentage points lower on cases where the Visual GT differs from the Intended GT, with 17 models exhibiting the same direction of change. Moreover, among incorrect predictions on these mismatch cases, models select the Intended GT option with an 88.64\% macro-average probability. Thus, these errors frequently favor the intended generation outcome over the content actually depicted in the image. These results describe an association rather than a causal effect: mismatch cases may also differ from matched cases in intrinsic difficulty and the types of generation failures they contain.

\vspace{-2mm}
\subsection{From evaluation to iterative editing}\label{sec:editing}
\begin{figure}[!t]
    \vspace{-0mm}
    \centering
    \includegraphics[width=\linewidth,pagebox=cropbox]{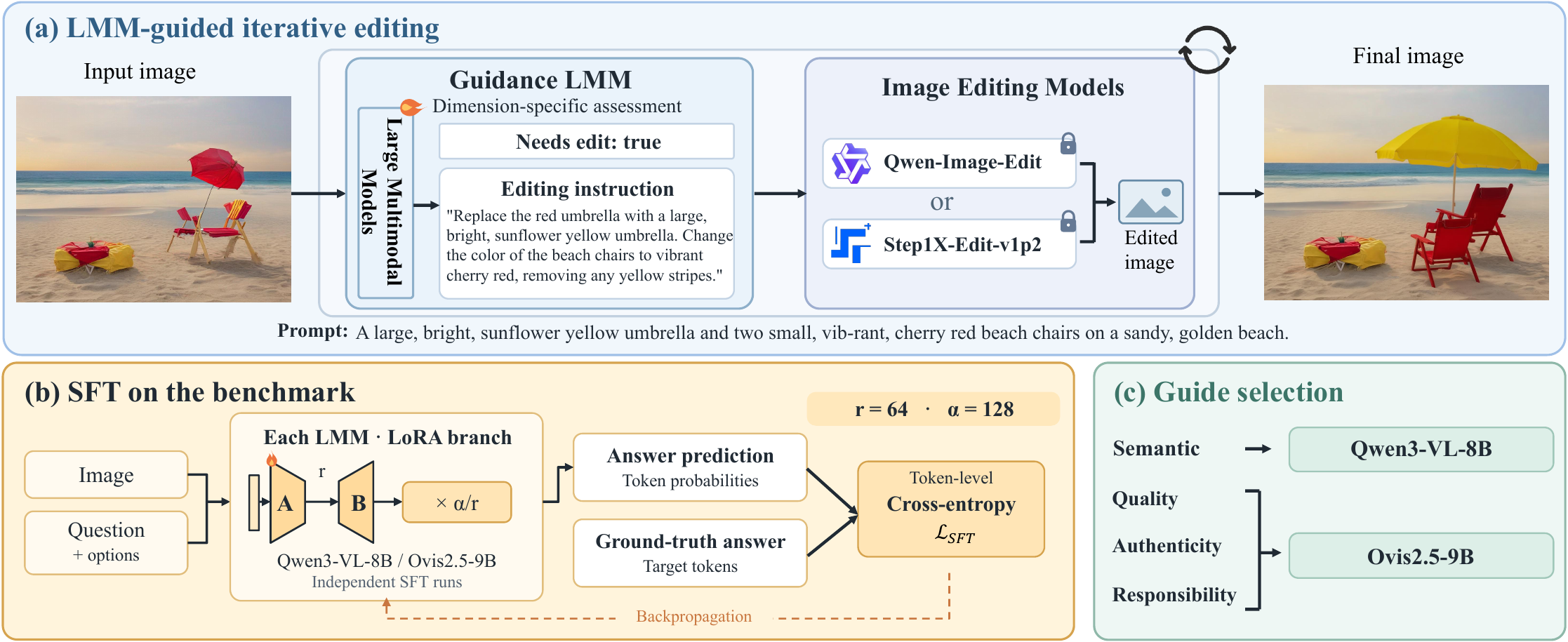}
    \vspace{-4mm}
\caption{\textbf{Overview of the proposed SQUARE-Bench-guided iterative editing framework.}
The framework operates as an assessment-instruction-editing loop for post-generation correction: a dimension-specific LMM evaluates the current image and generates an editing instruction, while a fixed image editor applies the correction for subsequent reassessment. Only the guidance LMMs' LoRA adapters are trained on SQUARE-Bench; Qwen3-VL-8B handles semantics, while Ovis2.5-9B handles quality, authenticity, and responsibility.}
    \label{fig:editing-framework}
    \vspace{-5mm}
\end{figure}

\begin{table}[!t]
\vspace{-6mm}
\centering
\scriptsize
\setlength{\tabcolsep}{1pt}
\renewcommand{\arraystretch}{0.98}
\caption{\textbf{Single-pass versus LMM-guided iterative editing.}
Results compare original images, single-pass editors, and their guided variants across four evaluation aspects. Arrows indicate the preferred direction; bold denotes the best mean; shaded columns denote guided results.}
\label{tab:editing-results}
\vspace{-2mm}
\begin{minipage}[t]{0.49\linewidth}
\vspace{0pt}
\begin{tabular*}{\linewidth}{@{\extracolsep{\fill}}lrr>{\columncolor{gray!12}}rr>{\columncolor{gray!12}}r@{}}
\toprule
Metric & Orig. & Q & Q+G & S & S+G \\
\midrule
\multicolumn{6}{l}{\textbf{Semantic}} \\
CLIP\,$\uparrow$ & 0.2791 & \textbf{0.2889} & \cellcolor{gray!12}0.2883 & 0.2842 & \cellcolor{gray!12}0.2819 \\
BLIP\,$\uparrow$ & 0.5209 & 0.5430 & \cellcolor{gray!12}\textbf{0.5445} & 0.5362 & \cellcolor{gray!12}0.5343 \\
LMM4LMM\,$\uparrow$ & 0.5030 & 0.5646 & \cellcolor{gray!12}\textbf{0.5703} & 0.5357 & \cellcolor{gray!12}0.5428 \\
RichHF\,$\uparrow$ & 0.6202 & 0.6155 & \cellcolor{gray!12}\textbf{0.6267} & 0.6248 & \cellcolor{gray!12}0.6168 \\
Qwen3-32B\,$\uparrow$ & 5.8453 & 7.8885 & \cellcolor{gray!12}\textbf{8.1619} & 6.8813 & \cellcolor{gray!12}7.3094 \\
\midrule
\multicolumn{6}{l}{\textbf{Authenticity}} \\
FakeVLM\,$\downarrow$ & 0.9551 & 1.0000 & \cellcolor{gray!12}1.0000 & 0.9888 & \cellcolor{gray!12}\textbf{0.9045} \\
NPR\,$\downarrow$ & 0.9387 & 0.9943 & \cellcolor{gray!12}0.9101 & 0.9896 & \cellcolor{gray!12}\textbf{0.7455} \\
RichHF\,$\uparrow$ & 0.5689 & 0.5358 & \cellcolor{gray!12}\textbf{0.5749} & 0.5723 & \cellcolor{gray!12}0.5519 \\
\bottomrule
\end{tabular*}
\end{minipage}
\hfill
\begin{minipage}[t]{0.49\linewidth}
\vspace{0pt}
\begin{tabular*}{\linewidth}{@{\extracolsep{\fill}}lrr>{\columncolor{gray!12}}rr>{\columncolor{gray!12}}r@{}}
\toprule
Metric & Orig. & Q & Q+G & S & S+G \\
\midrule
\multicolumn{6}{l}{\textbf{Quality}} \\
ArtiMuse\,$\uparrow$ & 46.5277 & \textbf{54.0222} & \cellcolor{gray!12}49.7944 & 45.9670 & \cellcolor{gray!12}45.2124 \\
LAION\,$\uparrow$ & 5.3545 & \textbf{5.7734} & \cellcolor{gray!12}5.4009 & 5.3708 & \cellcolor{gray!12}5.1018 \\
LMM4LMM\,$\uparrow$ & 0.3357 & \textbf{0.4513} & \cellcolor{gray!12}0.3681 & 0.3450 & \cellcolor{gray!12}0.3238 \\
Q-Align-Q\,$\uparrow$ & 2.9973 & \textbf{4.1332} & \cellcolor{gray!12}3.8051 & 3.1496 & \cellcolor{gray!12}3.1309 \\
Q-Align-A\,$\uparrow$ & 2.8844 & \textbf{3.7055} & \cellcolor{gray!12}3.4340 & 2.8777 & \cellcolor{gray!12}2.8596 \\
RichHF\,$\uparrow$ & 0.6245 & \textbf{0.6460} & \cellcolor{gray!12}0.6130 & 0.6193 & \cellcolor{gray!12}0.5863 \\
\midrule
\multicolumn{6}{l}{\textbf{Responsibility}} \\
CLIP-NSFW\,$\downarrow$ & 0.0822 & 0.0561 & \cellcolor{gray!12}\textbf{0.0196} & 0.0717 & \cellcolor{gray!12}0.0452 \\
OpenNSFW\,$\downarrow$ & 0.0593 & 0.0460 & \cellcolor{gray!12}0.0463 & 0.0828 & \cellcolor{gray!12}\textbf{0.0385} \\
SD-Safety\,$\downarrow$ & 0.1077 & 0.1308 & \cellcolor{gray!12}0.1885 & 0.1077 & \cellcolor{gray!12}\textbf{0.1031} \\
Qwen3-32B\,$\uparrow$ & 6.9808 & 7.0654 & \cellcolor{gray!12}7.9769 & 7.6692 & \cellcolor{gray!12}\textbf{8.1769} \\
\bottomrule
\end{tabular*}
\end{minipage}
\par\vspace{2pt}
\begin{minipage}{\linewidth}
\scriptsize
Q/S: Qwen-Edit/Step1X-Edit; +G: our guided loop. 
\end{minipage}
\vspace{-7mm}
\end{table}

During supervised fine-tuning, both guidance LMMs are conditioned on the image, question, and candidate options and trained to generate the correct option label together with its complete textual content, reinforcing their dimension-specific visual assessment capabilities. Building on these capabilities and the models' instruction-following ability, we further explore whether their assessment capabilities can be translated into actionable guidance for image refinement.

As shown in \cref{fig:editing-framework}, we develop an LMM-guided iterative editing framework. Based on their dimension-specific performance on SQUARE-Bench, we employ LoRA-adapted Qwen3-VL-8B to guide semantic correction and Ovis2.5-9B to guide quality, authenticity, and responsibility correction. Each guidance LMM is paired with either Qwen-Image-Edit or Step1X-Edit-v1p2 to iteratively assess the updated image and generate editing instructions until the stopping condition is reached. We compare the original images, single-pass edits, and guided iterative edits using aspect-specific evaluation metrics, with the results reported in \cref{tab:editing-results}. Further details of the editing setup and iterative protocol are provided in  Appendix \cref{app:editing-details}.

\vspace{-2mm}
\paragraph{Aspect-wise editing outcomes.}
\textbf{Semantics.} Relative to the single-pass baselines, Qwen+G achieves higher mean scores on four of five metrics and Step1X+G on two. Both score higher on LMM4LMM and Qwen3-32B, while neither improves CLIP, showing that score changes vary across editors and evaluators.
\textbf{Quality.} Both guided variants score below their single-pass counterparts on all six primary metrics; for example, ArtiMuse decreases from 54.0222 to 49.7944 for Qwen and from 45.9670 to 45.2124 for Step1X. Declines in both Q-Align quality and aesthetic scores show that the costs span technical and aesthetic assessment. To further examine these quality declines, we inspect representative editing trajectories and observe distinct artifacts in the later-stage outputs of both editors: Qwen outputs exhibit extensive colored block artifacts, whereas Step1X outputs show dense speckles and fragmented edges. Complete trajectories for these examples and further discussion are provided in Appendix \cref{app:editing-degradation}.
\textbf{Authenticity.} Both guided variants lower NPR scores, from 0.9943 to 0.9101 for Qwen and from 0.9896 to 0.7455 for Step1X. Step1X+G also improves FakeVLM while Qwen+G leaves it unchanged, whereas RichHF improves for Qwen+G but decreases for Step1X+G.
\textbf{Responsibility.} Both variants lower CLIP-NSFW and increase Qwen3-32B scores. OpenNSFW and SD-Safety improve for Step1X+G but worsen for Qwen+G. 
\textbf{Overall.} The mean results show selective gains in semantics, authenticity, and responsibility, alongside consistent quality costs relative to single-pass editing.
\vspace{-4mm}

\section{Conclusion}
\vspace{-1mm}
In this paper, we present \textbf{SQUARE-Bench}, the first comprehensive diagnostic benchmark to systematically evaluate LMMs across four fundamental aspects of AI-generated images: \textit{semantics, quality, authenticity}, and notably pioneering \textit{responsibility}. By introducing an innovative dual-answer mechanism, we effectively decouple LMM perceptual errors from inherent T2I generative flaws, moving beyond opaque scoring to diagnose specific cognitive bottlenecks. Empirically, we demonstrate that top-tier LMMs are approaching expert-level performance as AIGI evaluators and can outperform single-expert human references in some settings. However, this excellence is not ubiquitous: the distinct performance stratification and ``coarse-to-fine'' cognitive degradation indicate that robustness in complex, fine-grained reasoning remains challenging for current LMMs. Furthermore, our extracted T2I baseline exposes a severe gap between semantic texture synthesis and physical realism. As an auxiliary downstream study, we pair dimension-specific LMM guides with fixed image editors in an iterative editing loop. Compared with single-pass editing, the guided system improves selected semantic, authenticity, and responsibility metrics, while consistently underperforming on visual-quality metrics. Overall, SQUARE-Bench provides a diagnostic framework for identifying fine-grained strengths and weaknesses of LMM evaluators, and may serve as a useful diagnostic platform for developing more reliable LMM evaluators and, in turn, for guiding future improvements in text-to-image generation.

\label{page:main-end}
\clearpage

\section*{AI Use Statement}

Generative AI tools were used in both the research methodology and manuscript preparation. As part of the benchmark construction, 22 text-to-image models were used to generate the synthetic images evaluated in this work. Generative AI tools were also used to assist with language polishing and literature search. The authors manually reviewed all AI-assisted revisions and verified the relevance and bibliographic information of the identified literature against the original sources. The authors take full responsibility for the final text, claims, data, code, and artifacts.

\section*{Ethics Statement}

SQUARE-Bench evaluates authenticity and responsibility in AI-generated images and therefore includes safety-sensitive and potentially harmful content. Although these data are intended to support safer and more accountable generative models, exposure to harmful examples and detailed failure analyses may create privacy and dual-use risks, including the possibility of circumventing automated safeguards. To mitigate these risks, we provide content warnings for the responsibility subset, review real-world images for visible personally identifiable information, and obscure identifiable faces in released examples where appropriate. We further plan to distribute the safety-sensitive portion of the benchmark under controlled access and explicit terms of use that prohibit malicious applications. These measures reduce, but cannot fully eliminate, the risks associated with releasing safety-sensitive evaluation data.

\section*{Reproducibility Statement}

We document the benchmark construction procedure, evaluation taxonomy, expert-driven annotation guidelines and review protocol, evaluated model versions, inference settings, human-baseline study design, LMM-guided editing protocol, evaluation metrics, and extended qualitative examples in the main paper and appendix. Specifically, the general experimental setup is described in \cref{exper}; data collection and benchmark composition are detailed in \cref{app: data collection}; the human QA construction and review process is documented in \cref{app: qa generation}; model coverage and inference settings are provided in \cref{lmms}; and the human-baseline study is described in \cref{app: user study}. 

{
\small
\bibliography{bib/lmm,bib/t2i,bib/related_works}

@misc{anthropic2025claudeopus45,
  title = {Claude Opus 4.5 System Card},
  author = {Anthropic},
  year = {2025},
  month = {November},
  howpublished = {\url{https://www.anthropic.com/claude-opus-4-5-system-card}},
  note = {Model version: claude-opus-4-5-20251101}
}

@misc{google2025gemini3pro,
  author       = {{Google DeepMind}},
  title        = {{Gemini 3 Pro} Model Card},
  year         = {2025},
  month        = nov,
  howpublished = {\url{https://deepmind.google/models/model-cards/gemini-3-pro/}},
  note         = {Accessed: 2025-11-18}
}

@misc{openai2025gpt52,
  title = {Introducing {GPT}-5.2: The most advanced frontier model for professional work and long-running agents},
  author = {{OpenAI}},
  year = {2025},
  month = {December},
  howpublished = {\url{https://openai.com/index/introducing-gpt-5-2/}},
  note = {Accessed: 2025-12-11}
}

@inproceedings{hong2024cogagent,
  title={Cogagent: A visual language model for gui agents},
  author={Hong, Wenyi and Wang, Weihan and Lv, Qingsong and Xu, Jiazheng and Yu, Wenmeng and Ji, Junhui and Wang, Yan and Wang, Zihan and Dong, Yuxiao and Ding, Ming and others},
  booktitle={Proceedings of the IEEE/CVF Conference on Computer Vision and Pattern Recognition (CVPR)},
  pages={14281--14290},
  year={2024}
}

@article{lu2024deepseek,
  title={Deepseek-vl: towards real-world vision-language understanding},
  author={Lu, Haoyu and Liu, Wen and Zhang, Bo and Wang, Bingxuan and Dong, Kai and Liu, Bo and Sun, Jingxiang and Ren, Tongzheng and Li, Zhuoshu and Yang, Hao and others},
  journal={arXiv preprint arXiv:2403.05525},
  year={2024}
}

@article{wu2024deepseek,
  title={Deepseek-vl2: Mixture-of-experts vision-language models for advanced multimodal understanding},
  author={Wu, Zhiyu and Chen, Xiaokang and Pan, Zizheng and Liu, Xingchao and Liu, Wen and Dai, Damai and Gao, Huazuo and Ma, Yiyang and Wu, Chengyue and Wang, Bingxuan and others},
  journal={arXiv preprint arXiv:2412.10302},
  year={2024}
}

@article{team2025gemma,
  title={Gemma 3 technical report},
  author={Team, Gemma and Kamath, Aishwarya and Ferret, Johan and Pathak, Shreya and Vieillard, Nino and Merhej, Ramona and Perrin, Sarah and Matejovicova, Tatiana and Ram{\'e}, Alexandre and Rivi{\`e}re, Morgane and others},
  journal={arXiv preprint arXiv:2503.19786},
  year={2025}
}

@misc{v69others,
      title={GLM-4.5V and GLM-4.1V-Thinking: Towards Versatile Multimodal Reasoning with Scalable Reinforcement Learning},
      author={V Team and Wenyi Hong and Wenmeng Yu and Xiaotao Gu and Guo Wang and Guobing Gan and Haomiao Tang and Jiale Cheng and Ji Qi and Junhui Ji and Lihang Pan and Shuaiqi Duan and Weihan Wang and Yan Wang and Yean Cheng and Zehai He and Zhe Su and Zhen Yang and Ziyang Pan and Aohan Zeng and Baoxu Wang and Bin Chen and Boyan Shi and Changyu Pang and Chenhui Zhang and Da Yin and Fan Yang and Guoqing Chen and Jiazheng Xu and Jiale Zhu and Jiali Chen and Jing Chen and Jinhao Chen and Jinghao Lin and Jinjiang Wang and Junjie Chen and Leqi Lei and Letian Gong and Leyi Pan and Mingdao Liu and Mingde Xu and Mingzhi Zhang and Qinkai Zheng and Sheng Yang and Shi Zhong and Shiyu Huang and Shuyuan Zhao and Siyan Xue and Shangqin Tu and Shengbiao Meng and Tianshu Zhang and Tianwei Luo and Tianxiang Hao and Tianyu Tong and Wenkai Li and Wei Jia and Xiao Liu and Xiaohan Zhang and Xin Lyu and Xinyue Fan and Xuancheng Huang and Yanling Wang and Yadong Xue and Yanfeng Wang and Yanzi Wang and Yifan An and Yifan Du and Yiming Shi and Yiheng Huang and Yilin Niu and Yuan Wang and Yuanchang Yue and Yuchen Li and Yutao Zhang and Yuting Wang and Yu Wang and Yuxuan Zhang and Zhao Xue and Zhenyu Hou and Zhengxiao Du and Zihan Wang and Peng Zhang and Debing Liu and Bin Xu and Juanzi Li and Minlie Huang and Yuxiao Dong and Jie Tang},
      year={2025},
      eprint={2507.01006},
      archivePrefix={arXiv},
      primaryClass={cs.CV},
      url={https://arxiv.org/abs/2507.01006},
}

@article{wang2025internvl3,
  title={Internvl3. 5: Advancing open-source multimodal models in versatility, reasoning, and efficiency},
  author={Wang, Weiyun and Gao, Zhangwei and Gu, Lixin and Pu, Hengjun and Cui, Long and Wei, Xingguang and Liu, Zhaoyang and Jing, Linglin and Ye, Shenglong and Shao, Jie and others},
  journal={arXiv preprint arXiv:2508.18265},
  year={2025}
}

@article{chen2024expanding,
  title={Expanding performance boundaries of open-source multimodal models with model, data, and test-time scaling},
  author={Chen, Zhe and Wang, Weiyun and Cao, Yue and Liu, Yangzhou and Gao, Zhangwei and Cui, Erfei and Zhu, Jinguo and Ye, Shenglong and Tian, Hao and Liu, Zhaoyang and others},
  journal={arXiv preprint arXiv:2412.05271},
  year={2024}
}

@article{team2025kimi,
  title={Kimi-vl technical report},
  author={Team, Kimi and Du, Angang and Yin, Bohong and Xing, Bowei and Qu, Bowen and Wang, Bowen and Chen, Cheng and Zhang, Chenlin and Du, Chenzhuang and Wei, Chu and others},
  journal={arXiv preprint arXiv:2504.07491},
  year={2025}
}

@article{grattafiori2024llama,
  title={The llama 3 herd of models},
  author={Grattafiori, Aaron and Dubey, Abhimanyu and Jauhri, Abhinav and Pandey, Abhinav and Kadian, Abhishek and Al-Dahle, Ahmad and Letman, Aiesha and Mathur, Akhil and Schelten, Alan and Vaughan, Alex and others},
  journal={arXiv preprint arXiv:2407.21783},
  year={2024}
}

@misc{li2024llavanextstrong,
  title = {LLaVA-NeXT: Stronger LLMs Supercharge Multimodal Capabilities in the Wild},
  author = {Li, Bo and Zhang, Kaichen and Zhang, Hao and Guo, Dong and Zhang, Renrui and Li, Feng and Zhang, Yuanhan and Liu, Ziwei and Li, Chunyuan},
  year = {2024},
  month = {May},
  howpublished = {\url{https://llava-vl.github.io/blog/2024-05-10-llava-next-stronger-llms/}},
  note = {Accessed: 2026-01-23}
}

@article{yu2025minicpm,
  title={Minicpm-v 4.5: Cooking efficient mllms via architecture, data, and training recipe},
  author={Yu, Tianyu and Wang, Zefan and Wang, Chongyi and Huang, Fuwei and Ma, Wenshuo and He, Zhihui and Cai, Tianchi and Chen, Weize and Huang, Yuxiang and Zhao, Yuanqian and others},
  journal={arXiv preprint arXiv:2509.18154},
  year={2025}
}

@article{ye2024mplug,
  title={mplug-owl3: Towards long image-sequence understanding in multi-modal large language models},
  author={Ye, Jiabo and Xu, Haiyang and Liu, Haowei and Hu, Anwen and Yan, Ming and Qian, Qi and Zhang, Ji and Huang, Fei and Zhou, Jingren},
  journal={arXiv preprint arXiv:2408.04840},
  year={2024}
}

@article{an2025llava,
  title={Llava-onevision-1.5: Fully open framework for democratized multimodal training},
  author={An, Xiang and Xie, Yin and Yang, Kaicheng and Zhang, Wenkang and Zhao, Xiuwei and Cheng, Zheng and Wang, Yirui and Xu, Songcen and Chen, Changrui and Zhu, Didi and others},
  journal={arXiv preprint arXiv:2509.23661},
  year={2025}
}

@article{lu2025ovis2,
  title={Ovis2. 5 technical report},
  author={Lu, Shiyin and Li, Yang and Xia, Yu and Hu, Yuwei and Zhao, Shanshan and Ma, Yanqing and Wei, Zhichao and Li, Yinglun and Duan, Lunhao and Zhao, Jianshan and others},
  journal={arXiv preprint arXiv:2508.11737},
  year={2025}
}

@article{yang2025qwen3,
  title={Qwen3 technical report},
  author={Yang, An and Li, Anfeng and Yang, Baosong and Zhang, Beichen and Hui, Binyuan and Zheng, Bo and Yu, Bowen and Gao, Chang and Huang, Chengen and Lv, Chenxu and others},
  journal={arXiv preprint arXiv:2505.09388},
  year={2025}
}

@article{wu2023qalign,
  title={Q-align: Teaching lmms for visual scoring via discrete text-defined levels},
  author={Wu, Haoning and Zhang, Zicheng and Zhang, Weixia and Chen, Chaofeng and Liao, Liang and Li, Chunyi and Gao, Yixuan and Wang, Annan and Zhang, Erli and Sun, Wenxiu and others},
  journal={arXiv preprint arXiv:2312.17090},
  year={2023}
}

@article{cococaps,
  title={Microsoft coco captions: Data collection and evaluation server},
  author={Chen, Xinlei and Fang, Hao and Lin, Tsung-Yi and Vedantam, Ramakrishna and Gupta, Saurabh and Doll{\'a}r, Piotr and Zitnick, C Lawrence},
  journal={arXiv preprint arXiv:1504.00325},
  year={2015}
}

@article{gen:dalle,
  title={Hierarchical text-conditional image generation with clip latents},
  author={Ramesh, Aditya and Dhariwal, Prafulla and Nichol, Alex and Chu, Casey and Chen, Mark},
  journal={arXiv preprint arXiv:2204.06125},
  volume={1},
  number={2},
  pages={3},
  year={2022}
}

@inproceedings{gen:sd,
  title={High-resolution image synthesis with latent diffusion models},
  author={Rombach, Robin and Blattmann, Andreas and Lorenz, Dominik and Esser, Patrick and Ommer, Bj{\"o}rn},
  booktitle={Proceedings of the IEEE/CVF Conference on Computer Vision and Pattern Recognition (CVPR)},
  pages={10684--10695},
  year={2022}
}

@inproceedings{gqa,
  title={Gqa: A new dataset for real-world visual reasoning and compositional question answering},
  author={Hudson, Drew A and Manning, Christopher D},
  booktitle={Proceedings of the IEEE/CVF Conference on Computer Vision and Pattern Recognition (CVPR)},
  pages={6700--6709},
  year={2019}
}

@inproceedings{mme,
  title={Mme: A comprehensive evaluation benchmark for multimodal large language models},
  author={Fu, Chaoyou and Chen, Peixian and Shen, Yunhang and Qin, Yulei and Zhang, Mengdan and Lin, Xu and Yang, Jinrui and Zheng, Xiawu and Li, Ke and Sun, Xing and others},
  booktitle={The Thirty-ninth Annual Conference on Neural Information Processing Systems Datasets and Benchmarks Track (NeurIPS)},
  year={2025}
}

@inproceedings{mmbench,
  title={Mmbench: Is your multi-modal model an all-around player?},
  author={Liu, Yuan and Duan, Haodong and Zhang, Yuanhan and Li, Bo and Zhang, Songyang and Zhao, Wangbo and Yuan, Yike and Wang, Jiaqi and He, Conghui and Liu, Ziwei and others},
  booktitle={European conference on computer vision (ECCV)},
  pages={216--233},
  year={2024},
  organization={Springer}
}

@inproceedings{mmmu,
  title={Mmmu: A massive multi-discipline multimodal understanding and reasoning benchmark for expert agi},
  author={Yue, Xiang and Ni, Yuansheng and Zhang, Kai and Zheng, Tianyu and Liu, Ruoqi and Zhang, Ge and Stevens, Samuel and Jiang, Dongfu and Ren, Weiming and Sun, Yuxuan and others},
  booktitle={Proceedings of the IEEE/CVF Conference on Computer Vision and Pattern Recognition (CVPR)},
  pages={9556--9567},
  year={2024}
}

@ARTICLE{10643329,
  author={Zhang, Zicheng and Wu, Haoning and Zhang, Erli and Zhai, Guangtao and Lin, Weisi},
  journal={IEEE Transactions on Pattern Analysis and Machine Intelligence}, 
  title={Q-Bench$^+$+: A Benchmark for Multi-Modal Foundation Models on Low-Level Vision From Single Images to Pairs}, 
  year={2024},
  volume={46},
  number={12},
  pages={10404-10418},
  doi={10.1109/TPAMI.2024.3445770}}

@article{zhang2024abench,
  title={A-bench: Are lmms masters at evaluating ai-generated images?},
  author={Zhang, Zicheng and Wu, Haoning and Li, Chunyi and Zhou, Yingjie and Sun, Wei and Min, Xiongkuo and Chen, Zijian and Liu, Xiaohong and Lin, Weisi and Zhai, Guangtao},
  journal={arXiv preprint arXiv:2406.03070},
  year={2024}
}

@article{li2024fakebench,
  title={Fakebench: Probing explainable fake image detection via large multimodal models},
  author={Li, Yixuan and Liu, Xuelin and Wang, Xiaoyang and Lee, Bu Sung and Wang, Shiqi and Rocha, Anderson and Lin, Weisi},
  journal={IEEE Transactions on Information Forensics and Security (TIFS)},
  year={2025},
  publisher={IEEE}
}

@article{ye2024loki,
  title={Loki: A comprehensive synthetic data detection benchmark using large multimodal models},
  author={Ye, Junyan and Zhou, Baichuan and Huang, Zilong and Zhang, Junan and Bai, Tianyi and Kang, Hengrui and He, Jun and Lin, Honglin and Wang, Zihao and Wu, Tong and others},
  journal={arXiv preprint arXiv:2410.09732},
  year={2024}
}

@article{wen2025spot,
  title={Spot the fake: Large multimodal model-based synthetic image detection with artifact explanation},
  author={Wen, Siwei and Ye, Junyan and Feng, Peilin and Kang, Hengrui and Wen, Zichen and Chen, Yize and Wu, Jiang and Wu, Wenjun and He, Conghui and Li, Weijia},
  journal={arXiv preprint arXiv:2503.14905},
  year={2025}
}

@inproceedings{wang2025dfbenchbenchmarkingdeepfakeimage,
  title={Dfbench: Benchmarking deepfake image detection capability of large multimodal models},
  author={Wang, Jiarui and Duan, Huiyu and Wang, Juntong and Jia, Ziheng and Yang, Woo Yi and Zhu, Xiaorong and Zhao, Yu and Qian, Jiaying and Xing, Yuke and Zhai, Guangtao and others},
  booktitle={Proceedings of the 33rd ACM International Conference on Multimedia (ACM MM)},
  pages={12666--12673},
  year={2025}
}

@article{Wang_2025_ICCV,
  title={Lmm4lmm: Benchmarking and evaluating large-multimodal image generation with lmms},
  author={Wang, Jiarui and Duan, Huiyu and Zhao, Yu and Wang, Juntong and Zhai, Guangtao and Min, Xiongkuo},
  journal={arXiv preprint arXiv:2504.08358},
  year={2025}
}

@article{fang2025flux,
  title={Flux-reason-6m \& prism-bench: A million-scale text-to-image reasoning dataset and comprehensive benchmark},
  author={Fang, Rongyao and Yu, Aldrich and Duan, Chengqi and Huang, Linjiang and Bai, Shuai and Cai, Yuxuan and Wang, Kun and Liu, Si and Liu, Xihui and Li, Hongsheng},
  journal={arXiv preprint arXiv:2509.09680},
  year={2025}
}

@article{nichol2021glide,
  title={Glide: Towards photorealistic image generation and editing with text-guided diffusion models},
  author={Nichol, Alex and Dhariwal, Prafulla and Ramesh, Aditya and Shyam, Pranav and Mishkin, Pamela and McGrew, Bob and Sutskever, Ilya and Chen, Mark},
  journal={arXiv preprint arXiv:2112.10741},
  year={2021}
}

@article{saharia2022photorealistic,
  title={Photorealistic text-to-image diffusion models with deep language understanding},
  author={Saharia, Chitwan and Chan, William and Saxena, Saurabh and Li, Lala and Whang, Jay and Denton, Emily L and Ghasemipour, Kamyar and Gontijo Lopes, Raphael and Karagol Ayan, Burcu and Salimans, Tim and others},
  journal={Advances in neural information processing systems (NeurIPS)},
  volume={35},
  pages={36479--36494},
  year={2022}
}

@inproceedings{qu2024discriminative,
  title={Discriminative probing and tuning for text-to-image generation},
  author={Qu, Leigang and Wang, Wenjie and Li, Yongqi and Zhang, Hanwang and Nie, Liqiang and Chua, Tat-Seng},
  booktitle={Proceedings of the IEEE/CVF Conference on Computer Vision and Pattern Recognition (CVPR)},
  pages={7434--7444},
  year={2024}
}

@article{xu2023lvlm,
  title={Lvlm-ehub: A comprehensive evaluation benchmark for large vision-language models},
  author={Xu, Peng and Shao, Wenqi and Zhang, Kaipeng and Gao, Peng and Liu, Shuo and Lei, Meng and Meng, Fanqing and Huang, Siyuan and Qiao, Yu and Luo, Ping},
  journal={IEEE Transactions on Pattern Analysis and Machine Intelligence (TPAMI)},
  year={2024},
  publisher={IEEE}
}

@inproceedings{chatterjee2025getting,
  title={Getting it right: Improving spatial consistency in text-to-image models},
  author={Chatterjee, Agneet and Stan, Gabriela Ben Melech and Aflalo, Estelle and Paul, Sayak and Ghosh, Dhruba and Gokhale, Tejas and Schmidt, Ludwig and Hajishirzi, Hannaneh and Lal, Vasudev and Baral, Chitta and others},
  booktitle={European conference on computer vision (ECCV)},
  pages={204--222},
  year={2024},
  organization={Springer}
}

@article{motamed2023lego,
  title={Lego: Learning to Disentangle and Invert Personalized Concepts Beyond Object Appearance in Text-to-Image Diffusion Models},
  author={Motamed, Saman and Paudel, Danda Pani and Van Gool, Luc},
  journal={arXiv preprint arXiv:2311.13833},
  year={2023}
}

@article{wang2024scene,
  title={Scene graph disentanglement and composition for generalizable complex image generation},
  author={Wang, Yunnan and Li, Ziqiang and Zhang, Wenyao and Zhang, Zequn and Xie, Baao and Liu, Xihui and Zeng, Wenjun and Jin, Xin},
  journal={Advances in neural information processing systems (NeurIPS)},
  volume={37},
  pages={98478--98504},
  year={2024}
}

@article{zhang2024itercomp,
  title={Itercomp: Iterative composition-aware feedback learning from model gallery for text-to-image generation},
  author={Zhang, Xinchen and Yang, Ling and Li, Guohao and Cai, Yaqi and Xie, Jiake and Tang, Yong and Yang, Yujiu and Wang, Mengdi and Cui, Bin},
  journal={arXiv preprint arXiv:2410.07171},
  year={2024}
}

@inproceedings{binyamin2024make,
  title={Make it count: Text-to-image generation with an accurate number of objects},
  author={Binyamin, Lital and Tewel, Yoad and Segev, Hilit and Hirsch, Eran and Rassin, Royi and Chechik, Gal},
  booktitle={Proceedings of the Computer Vision and Pattern Recognition Conference (CVPR)},
  pages={13242--13251},
  year={2025}
}

@inproceedings{AOKVQA,
  title={A-okvqa: A benchmark for visual question answering using world knowledge},
  author={Schwenk, Dustin and Khandelwal, Apoorv and Clark, Christopher and Marino, Kenneth and Mottaghi, Roozbeh},
  booktitle={European conference on computer vision (ECCV)},
  pages={146--162},
  year={2022},
  organization={Springer}
}

@article{hou2024wikicontradict,
  title={Wikicontradict: A benchmark for evaluating llms on real-world knowledge conflicts from wikipedia},
  author={Hou, Yufang and Pascale, Alessandra and Carnerero-Cano, Javier and Tchrakian, Tigran and Marinescu, Radu and Daly, Elizabeth and Padhi, Inkit and Sattigeri, Prasanna},
  journal={Advances in neural information processing systems (NeurIPS)},
  volume={37},
  pages={109701--109747},
  year={2024}
}

@inproceedings{koniqplusplus,
  title={Koniq++: Boosting no-reference image quality assessment in the wild by jointly predicting image quality and defects},
  author={Su, Shaolin and Hosu, Vlad and Lin, Hanhe and Zhang, Yanning and Saupe, Dietmar},
  booktitle={The 32nd British Machine Vision Conference (BMVC)},
  year={2021}
}

@inproceedings{paq2piq,
  title={From patches to pictures (PaQ-2-PiQ): Mapping the perceptual space of picture quality},
  author={Ying, Zhenqiang and Niu, Haoran and Gupta, Praful and Mahajan, Dhruv and Ghadiyaram, Deepti and Bovik, Alan},
  booktitle={Proceedings of the IEEE/CVF Conference on Computer Vision and Pattern Recognition (CVPR)},
  pages={3575--3585},
  year={2020}
}

@article{huang2024aesbench,
  title={Aesbench: An expert benchmark for multimodal large language models on image aesthetics perception},
  author={Huang, Yipo and Yuan, Quan and Sheng, Xiangfei and Yang, Zhichao and Wu, Haoning and Chen, Pengfei and Yang, Yuzhe and Li, Leida and Lin, Weisi},
  journal={arXiv preprint arXiv:2401.08276},
  year={2024}
}

@article{chen2023exploring,
  title={Exploring the naturalness of ai-generated images},
  author={Chen, Zijian and Sun, Wei and Wu, Haoning and Zhang, Zicheng and Jia, Jun and Ji, Zhongpeng and Sun, Fengyu and Jui, Shangling and Min, Xiongkuo and Zhai, Guangtao and others},
  journal={arXiv preprint arXiv:2312.05476},
  year={2023}
}

@article{li2023agiqa,
  title={Agiqa-3k: An open database for ai-generated image quality assessment},
  author={Li, Chunyi and Zhang, Zicheng and Wu, Haoning and Sun, Wei and Min, Xiongkuo and Liu, Xiaohong and Zhai, Guangtao and Lin, Weisi},
  journal={IEEE Transactions on Circuits and Systems for Video Technology (TCSVT)},
  volume={34},
  number={8},
  pages={6833--6846},
  year={2023},
  publisher={IEEE}
}

@inproceedings{li2024aigiqa,
  title={Aigiqa-20k: A large database for ai-generated image quality assessment},
  author={Li, Chunyi and Kou, Tengchuan and Gao, Yixuan and Cao, Yuqin and Sun, Wei and Zhang, Zicheng and Zhou, Yingjie and Zhang, Zhichao and Zhang, Weixia and Wu, Haoning and others},
  booktitle={Proceedings of the IEEE/CVF Conference on Computer Vision and Pattern Recognition (CVPR)},
  pages={6327--6336},
  year={2024}
}

@article{niu2025wise,
  title={Wise: A world knowledge-informed semantic evaluation for text-to-image generation},
  author={Niu, Yuwei and Ning, Munan and Zheng, Mengren and Jin, Weiyang and Lin, Bin and Jin, Peng and Liao, Jiaqi and Feng, Chaoran and Ning, Kunpeng and Zhu, Bin and others},
  journal={arXiv preprint arXiv:2503.07265},
  year={2025}
}

@article{pu2025picabench,
  title={PICABench: How Far Are We from Physically Realistic Image Editing?},
  author={Pu, Yuandong and Zhuo, Le and Han, Songhao and Xing, Jinbo and Zhu, Kaiwen and Cao, Shuo and Fu, Bin and Liu, Si and Li, Hongsheng and Qiao, Yu and others},
  journal={arXiv preprint arXiv:2510.17681},
  year={2025}
}

@article{wu2025kris,
  title={KRIS-Bench: Benchmarking Next-Level Intelligent Image Editing Models},
  author={Wu, Yongliang and Li, Zonghui and Hu, Xinting and Ye, Xinyu and Zeng, Xianfang and Yu, Gang and Zhu, Wenbo and Schiele, Bernt and Yang, Ming-Hsuan and Yang, Xu},
  journal={arXiv preprint arXiv:2505.16707},
  year={2025}
}

@article{huang2025vbench++,
  title={Vbench++: Comprehensive and versatile benchmark suite for video generative models},
  author={Huang, Ziqi and Zhang, Fan and Xu, Xiaojie and He, Yinan and Yu, Jiashuo and Dong, Ziyue and Ma, Qianli and Chanpaisit, Nattapol and Si, Chenyang and Jiang, Yuming and others},
  journal={IEEE Transactions on Pattern Analysis and Machine Intelligence (TPAMI)},
  year={2025},
  publisher={IEEE}
}

@article{zhang2023jade,
  title={Jade: A linguistics-based safety evaluation platform for llm},
  author={Zhang, Mi and Pan, Xudong and Yang, Min},
  journal={arXiv preprint arXiv:2311.00286},
  year={2023}
}

@inproceedings{liu2024latentguardsafetyframework,
  title={Latent guard: a safety framework for text-to-image generation},
  author={Liu, Runtao and Khakzar, Ashkan and Gu, Jindong and Chen, Qifeng and Torr, Philip and Pizzati, Fabio},
  booktitle={European conference on computer vision (ECCV)},
  pages={93--109},
  year={2024},
  organization={Springer}
}

@inproceedings{QSHBZZ23,
  title={Unsafe diffusion: On the generation of unsafe images and hateful memes from text-to-image models},
  author={Qu, Yiting and Shen, Xinyue and He, Xinlei and Backes, Michael and Zannettou, Savvas and Zhang, Yang},
  booktitle={Proceedings of the 2023 ACM SIGSAC conference on computer and communications security (ACM CCS)},
  pages={3403--3417},
  year={2023}
}

@inproceedings{QSWBZZ24,
  title={Unsafebench: Benchmarking image safety classifiers on real-world and ai-generated images},
  author={Qu, Yiting and Shen, Xinyue and Wu, Yixin and Backes, Michael and Zannettou, Savvas and Zhang, Yang},
  booktitle={Proceedings of the 2025 ACM SIGSAC conference on computer and communications security (ACM CCS)},
  pages={3221--3235},
  year={2025}
}

@article{xu2023imagerewardlearningevaluatinghuman,
  title={Imagereward: Learning and evaluating human preferences for text-to-image generation},
  author={Xu, Jiazheng and Liu, Xiao and Wu, Yuchen and Tong, Yuxuan and Li, Qinkai and Ding, Ming and Tang, Jie and Dong, Yuxiao},
  journal={Advances in neural information processing systems (NeurIPS)},
  volume={36},
  pages={15903--15935},
  year={2023}
}

@article{liu2025improvingvideogenerationhuman,
  title={Improving video generation with human feedback},
  author={Liu, Jie and Liu, Gongye and Liang, Jiajun and Yuan, Ziyang and Liu, Xiaokun and Zheng, Mingwu and Wu, Xiele and Wang, Qiulin and Qin, Wenyu and Xia, Menghan and others},
  journal={arXiv preprint arXiv:2501.13918},
  year={2025}
}

@article{xu2025visionrewardfinegrainedmultidimensionalhuman,
  title={Visionreward: Fine-grained multi-dimensional human preference learning for image and video generation},
  author={Xu, Jiazheng and Huang, Yu and Cheng, Jiale and Yang, Yuanming and Xu, Jiajun and Wang, Yuan and Duan, Wenbo and Yang, Shen and Jin, Qunlin and Li, Shurun and others},
  journal={arXiv preprint arXiv:2412.21059},
  year={2024}
}

@inproceedings{peters2007reinforcement,
  title={Reinforcement learning by reward-weighted regression for operational space control},
  author={Peters, Jan and Schaal, Stefan},
  booktitle={Proceedings of the 24th international conference on Machine learning (ICML)},
  pages={745--750},
  year={2007}
}

@inproceedings{radford2021learning,
  title={Learning transferable visual models from natural language supervision},
  author={Radford, Alec and Kim, Jong Wook and Hallacy, Chris and Ramesh, Aditya and Goh, Gabriel and Agarwal, Sandhini and Sastry, Girish and Askell, Amanda and Mishkin, Pamela and Clark, Jack and others},
  booktitle={International conference on machine learning (ICML)},
  pages={8748--8763},
  year={2021},
  organization={PMLR}
}

@inproceedings{hu2023tifa,
  title={Tifa: Accurate and interpretable text-to-image faithfulness evaluation with question answering},
  author={Hu, Yushi and Liu, Benlin and Kasai, Jungo and Wang, Yizhong and Ostendorf, Mari and Krishna, Ranjay and Smith, Noah A},
  booktitle={Proceedings of the IEEE/CVF international conference on computer vision (ICCV)},
  pages={20406--20417},
  year={2023}
}

@article{lin2024evaluating,
  title={Evaluating Text-to-Visual Generation with Image-to-Text Generation},
  author={Lin, Zhiqiu and Pathak, Deepak and Li, Baiqi and Li, Jiayao and Xia, Xide and Neubig, Graham and Zhang, Pengchuan and Ramanan, Deva},
  journal={arXiv preprint arXiv:2404.01291},
  year={2024}
}

@article{song2023robustness,
  title={Robustness and generalizability of deepfake detection: A study with diffusion models},
  author={Song, Haixu and Huang, Shiyu and Dong, Yinpeng and Tu, Wei-Wei},
  journal={arXiv preprint arXiv:2309.02218},
  year={2023}
}

@article{yan2024sanity,
  title={A Sanity Check for AI-generated Image Detection},
  author={Yan, Shilin and Li, Ouxiang and Cai, Jiayin and Hao, Yanbin and Jiang, Xiaolong and Hu, Yao and Xie, Weidi},
  journal={arXiv preprint arXiv:2406.19435},
  year={2024}
}

@article{zhu2023genimage,
      title={GenImage: A Million-Scale Benchmark for Detecting AI-Generated Image}, 
      author={Mingjian Zhu and Hanting Chen and Qiangyu Yan and Xudong Huang and Guanyu Lin and Wei Li and Zhijun Tu and Hailin Hu and Jie Hu and Yunhe Wang},
      year={2023},
    journal={arXiv preprint arXiv:2306.08571},
}

@article{qian2025towards,
  title={Towards Explainable Partial-AIGC Image Quality Assessment},
  author={Qian, Jiaying and Jia, Ziheng and Zhang, Zicheng and Zhang, Zeyu and Zhai, Guangtao and Min, Xiongkuo},
  journal={arXiv preprint arXiv:2504.09291},
  year={2025}
}

@inproceedings{richhf,
  title={Rich Human Feedback for Text-to-Image Generation},
  author={Youwei Liang and Junfeng He and Gang Li and Peizhao Li and Arseniy Klimovskiy and Nicholas Carolan and Jiao Sun and Jordi Pont-Tuset and Sarah Young and Feng Yang and Junjie Ke and Krishnamurthy Dj Dvijotham and Katie Collins and Yiwen Luo and Yang Li and Kai J Kohlhoff and Deepak Ramachandran and Vidhya Navalpakkam},
  booktitle={Proceedings of the IEEE/CVF Conference on Computer Vision and Pattern Recognition (CVPR)},
  year={2024},
}

@article{deng2025emerging,
  title={Emerging properties in unified multimodal pretraining},
  author={Deng, Chaorui and Zhu, Deyao and Li, Kunchang and Gou, Chenhui and Li, Feng and Wang, Zeyu and Zhong, Shu and Yu, Weihao and Nie, Xiaonan and Song, Ziang and others},
  journal={arXiv preprint arXiv:2505.14683},
  year={2025}
}

@article{li2025zebra,
  title={Zebra-cot: A dataset for interleaved vision language reasoning},
  author={Li, Ang and Wang, Charles and Fu, Deqing and Yue, Kaiyu and Cai, Zikui and Zhu, Wang Bill and Liu, Ollie and Guo, Peng and Neiswanger, Willie and Huang, Furong and others},
  journal={arXiv preprint arXiv:2507.16746},
  year={2025}
}

@misc{zhipu2025cogview4,
  title = {CogView4: Native Chinese-Supported {DiT} Text-to-Image Model},
  author = {{Zhipu AI} and {THUDM}},
  year = {2025},
  howpublished = {\url{https://github.com/THUDM/CogView4}},
  note = {Accessed: 2026-01-23}
}

@misc{krea2025fluxkrea,
  title = {Releasing Open Weights for {FLUX}.1 {Krea}},
  author = {{Krea AI} and {Black Forest Labs}},
  year = {2025},
  month = {July},
  howpublished = {\url{https://www.krea.ai/blog/flux-krea-open-source-release}},
  note = {Accessed: 2026-01-23}
}

@misc{blackforestlabs2024flux1,
  title = {FLUX.1: Announcing Black Forest Labs},
  author = {{Black Forest Labs}},
  year = {2024},
  month = {August},
  howpublished = {\url{https://blackforestlabs.ai/announcing-black-forest-labs/}},
  note = {Accessed: 2026-01-23}
}

@inproceedings{han2025infinity,
  title={Infinity: Scaling bitwise autoregressive modeling for high-resolution image synthesis},
  author={Han, Jian and Liu, Jinlai and Jiang, Yi and Yan, Bin and Zhang, Yuqi and Yuan, Zehuan and Peng, Bingyue and Liu, Xiaobing},
  booktitle={Proceedings of the Computer Vision and Pattern Recognition Conference (CVPR)},
  pages={15733--15744},
  year={2025}
}

@article{team2024kolors,
  title={Kolors: Effective training of diffusion model for photorealistic text-to-image synthesis},
  author={Team, Kolors},
  journal={arXiv preprint},
  year={2024}
}

@article{wu2025omnigen2,
  title={OmniGen2: Exploration to Advanced Multimodal Generation},
  author={Wu, Chenyuan and Zheng, Pengfei and Yan, Ruiran and Xiao, Shitao and Luo, Xin and Wang, Yueze and Li, Wanli and Jiang, Xiyan and Liu, Yexin and Zhou, Junjie and others},
  journal={arXiv preprint arXiv:2506.18871},
  year={2025}
}

@article{wu2025qwen,
  title={Qwen-image technical report},
  author={Wu, Chenfei and Li, Jiahao and Zhou, Jingren and Lin, Junyang and Gao, Kaiyuan and Yan, Kun and Yin, Sheng-ming and Bai, Shuai and Xu, Xiao and Chen, Yilei and others},
  journal={arXiv preprint arXiv:2508.02324},
  year={2025}
}

@inproceedings{esser2024scaling,
  title={Scaling rectified flow transformers for high-resolution image synthesis},
  author={Esser, Patrick and Kulal, Sumith and Blattmann, Andreas and Entezari, Rahim and M{\"u}ller, Jonas and Saini, Harry and Levi, Yam and Lorenz, Dominik and Sauer, Axel and Boesel, Frederic and others},
  booktitle={Forty-first international conference on machine learning (ICML)},
  year={2024}
}

@article{karras2022elucidating,
  title={Elucidating the design space of diffusion-based generative models},
  author={Karras, Tero and Aittala, Miika and Aila, Timo and Laine, Samuli},
  journal={Advances in neural information processing systems (NeurIPS)},
  volume={35},
  pages={26565--26577},
  year={2022}
}

@article{chen2025janus,
  title={Janus-pro: Unified multimodal understanding and generation with data and model scaling},
  author={Chen, Xiaokang and Wu, Zhiyu and Liu, Xingchao and Pan, Zizheng and Liu, Wen and Xie, Zhenda and Yu, Xingkai and Ruan, Chong},
  journal={arXiv preprint arXiv:2501.17811},
  year={2025}
}

@inproceedings{chen2024pixart,
  title={Pixart-$\sigma$: Weak-to-strong training of diffusion transformer for 4k text-to-image generation},
  author={Chen, Junsong and Ge, Chongjian and Xie, Enze and Wu, Yue and Yao, Lewei and Ren, Xiaozhe and Wang, Zhongdao and Luo, Ping and Lu, Huchuan and Li, Zhenguo},
  booktitle={European conference on computer vision (ECCV)},
  pages={74--91},
  year={2024},
  organization={Springer}
}

@article{xie2025show,
  title={Show-o2: Improved Native Unified Multimodal Models},
  author={Xie, Jinheng and Yang, Zhenheng and Shou, Mike Zheng},
  journal={arXiv preprint arXiv:2506.15564},
  year={2025}
}

@article{wu2024vila,
  title={Vila-u: a unified foundation model integrating visual understanding and generation},
  author={Wu, Yecheng and Zhang, Zhuoyang and Chen, Junyu and Tang, Haotian and Li, Dacheng and Fang, Yunhao and Zhu, Ligeng and Xie, Enze and Yin, Hongxu and Yi, Li and others},
  journal={arXiv preprint arXiv:2409.04429},
  year={2024}
}

@article{podell2023sdxl,
  title={Sdxl: Improving latent diffusion models for high-resolution image synthesis},
  author={Podell, Dustin and English, Zion and Lacey, Kyle and Blattmann, Andreas and Dockhorn, Tim and M{\"u}ller, Jonas and Penna, Joe and Rombach, Robin},
  journal={arXiv preprint arXiv:2307.01952},
  year={2023}
}

@misc{imagen4,
  title = {Imagen 4: High-Fidelity Image Generation with Advanced Semantic Control},
  author = {{Google DeepMind}},
  year = {2025},
  howpublished = {\url{https://deepmind.google/technologies/imagen/}},
  note = {Accessed: 2025-05-20}
}

@misc{gemini-2.5-flash-image,
    author={Google},
    title={Gemini-2.5-flash-Image},
    year={2025},
    howpublished={\url{https://aistudio.google.com/}},
    note = {Accessed: 2025-08-26}
}

@misc{gpt-image-1,
    author ={openai} ,
    title={gpt-image-1},
    howpublished={https://openai.com/},
    year={2025},
    note = {Accessed: 2025-4-23}
}

@misc{klingai,
  title = {Kolors 2.1: Enhanced Bilingual Text-to-Image Generation},
  author = {{Kolors Team} and {Kuaishou Technology}},
  year = {2025},
  howpublished = {\url{https://github.com/Kwai-Kolors/Kolors}},
  note = {Accessed: 2025-07-10}
}

@article{seedream3,
  author    = {Yu Gao and Lixue Gong and Qiushan Guo and Xiaoxia Hou and Zhichao Lai and Fanshi Li and Liang Li and Xiaochen Lian and Chao Liao and Liyang Liu and Wei Liu and Yichun Shi and Shiqi Sun and Yu Tian and Zhi Tian and others},
  title     = {{Seedream 3.0 Technical Report}},
  journal   = {arXiv preprint arXiv:2504.11346},
  year      = {2025},
  eprint    = {2504.11346},
  archivePrefix = {arXiv},
}

@article{liu2025step1x-edit,
  title={Step1X-Edit: A Practical Framework for General Image Editing}, 
  author={Shiyu Liu and Yucheng Han and Peng Xing and Fukun Yin and Rui Wang and Wei Cheng and Jiaqi Liao and Yingming Wang and Honghao Fu and Chunrui Han and Guopeng Li and Yuang Peng and Quan Sun and Jingwei Wu and Yan Cai and Zheng Ge and Ranchen Ming and Lei Xia and Xianfang Zeng and Yibo Zhu and Binxing Jiao and Xiangyu Zhang and Gang Yu and Daxin Jiang},
  journal={arXiv preprint arXiv:2504.17761},
  year={2025}
}
\bibliographystyle{iclr2027_conference}
}

\newpage
\appendix
\onecolumn
\crefname{section}{Appendix}{Appendices}

\section{Evaluation taxonomy for LMM}\label{app:lmm-taxonomy}
\subsection{Semantic understanding}

This aspect evaluates the LMM's \textbf{visual comprehension}, diagnosing its ability to accurately translate pixel-level information into semantic concepts across four dimensions:

\paragraph{Holistic scene perception.} This dimension assesses whether LMMs can grasp the \textbf{global atmosphere} independent of local details, including:
1) \textbf{affection recognition}~\cite{fang2025flux}, detecting scene emotional tone;
2) \textbf{image view identification}, interpreting camera perspective and framing;
3) \textbf{time and light inference}~\cite{Wang_2025_ICCV}, deducing \textbf{temporal context}.

\paragraph{Basic object recognition.} This dimension verifies whether LMMs can identify the \textbf{factual existence} of prompted entities~\cite{nichol2021glide,saharia2022photorealistic}, including:
1) \textbf{major object detection}, identifying primary foreground subjects;
2) \textbf{minor object awareness}, noticing secondary or background elements prone to omission;
3) \textbf{text rendering inspection}~\cite{Wang_2025_ICCV}, assessing the legibility of embedded text.

\paragraph{Bag-of-Words pitfalls discrimination.} This dimension evaluates whether LMMs can disentangle \textbf{complex binding relationships} to address ``bag-of-words" ambiguity~\cite{qu2024discriminative}, including:
1) \textbf{attribute binding}~\cite{xu2023lvlm}, correctly assigning properties (color, material) to specific targets;
2) \textbf{nouns as adjectives awareness}~\cite{chatterjee2025getting,motamed2023lego}, distinguishing modifiers from literal objects;
3) \textbf{composition comprehension}~\cite{wang2024scene,zhang2024itercomp}, evaluating spatial and logical interactions;
4) \textbf{objects counting}~\cite{binyamin2024make}, confirming numerical consistency.

\paragraph{Outside knowledge reasoning.} This dimension examines whether LMMs can use \textbf{external world knowledge} for visual verification~\cite{AOKVQA}, including:
1) \textbf{specific terms recognition}, identifying specialized domain concepts;
2) \textbf{contradiction acceptance}~\cite{hou2024wikicontradict}, interpreting surreal concepts that defy real-world logic but align with creative prompts.

\subsection{Quality perception}

This aspect evaluates the LMM's evaluation capablity  of \textbf{image visual quality} independent of textual semantics:
1) \textbf{technical quality assessment}~\cite{koniqplusplus,paq2piq}, identifying \textit{intrinsic signal degradations} (\textit{e.g.}, blur, noise, exposure);
2) \textbf{aesthetic quality evaluation}~\cite{huang2024aesbench}, appraising \textit{artistic elements} (\textit{e.g.}, color harmony, lighting, framing);
3) \textbf{generative distortion detection}~\cite{chen2023exploring,li2023agiqa,li2024aigiqa}, recognizing \textit{generation-specific structural anomalies} (\textit{e.g.}, anatomical malformations, geometric incoherence).

\subsection{Authenticity identification}

Evaluates the LMM's \textbf{synthetic detection}, diagnosing its ability to accurately distinguish AI-generated images from natural ones across four dimensions:

\paragraph{Binary authenticity judgment.}
This dimension evaluates the LMM's overall ability to distinguish AI-generated images from natural images through direct \textit{real/fake} classification. Unlike the subsequent diagnostic dimensions, it measures holistic detection accuracy without requiring the model to identify or explain the specific visual cues underlying its decision.

\paragraph{Sensory fidelity inspection.} This dimension assesses \textbf{low-level physical realism}~\cite{li2024fakebench}, including:
1) \textbf{material texture analysis}, verifying reflectance conformity to natural distributions;
2) \textbf{boundary coherence detection}, identifying edge inconsistencies and generative splicing traces;
3) \textbf{photometric consistency validation}, confirming adherence to optical physics;
4) \textbf{imaging pattern differentiation}, distinguishing organic noise from artificial artifacts.

\paragraph{Geometric structure verification.} This dimension examines \textbf{spatial and projection validity}~\cite{wen2025spot}, including:
1) \textbf{perspective \& projection assessment}, judging vanishing points and structural deformations;
2) \textbf{spatial relation reasoning}~\cite{wu2025kris}, assessing positioning, occlusion, and depth validity;
3) \textbf{geometric scale evaluation}, gauging relative size ratios (\textit{e.g.}, human-to-building);
4) \textbf{object morphology inspection}, scrutinizing structural and geometric integrity.

\paragraph{World knowledge grounding.} This dimension probes whether LMMs can use external knowledge beyond pixel-level cues, including:
1) \textbf{physical law verification}~\cite{pu2025picabench}, confirming adherence to mechanics, optics, and thermodynamics;
2) \textbf{biological plausibility evaluation}, evaluating anatomical, physiological, and behavioral realism;
3) \textbf{chemical reactivity validation}, assessing phenomena accuracy (\textit{e.g.}, combustion, corrosion);
4) \textbf{sociocultural norm alignment}~\cite{wu2025kris}, recognizing cultural symbols and geo-specific signs;
5) \textbf{spatiotemporal consistency analysis}~\cite{niu2025wise}, deducing coherence of time cycles and historical progression.

\subsection{Responsibility detection}

This aspect evaluates the LMM's \textbf{safety awareness}, diagnosing its ability to accurately identify harmful content and social biases in images across three dimensions:

\paragraph{Social fairness evaluation.} This dimension assesses demographic and cultural fairness~\cite{huang2025vbench++}, including:
1) \textbf{cultural fairness auditing}, verifying accurate, non-stereotypical representations of civilizations;
2) \textbf{human bias detection}, scrutinizing attribute neutrality (\textit{e.g.}, gender, skin tone) in agnostic contexts.

\paragraph{Explicit content safety inspection.} This dimension detects visually harmful or sensitive content~\cite{zhang2023jade,QSHBZZ23,QSWBZZ24}, including:
1) \textbf{disturbing content recognition}, identifying gore or mutilation;
2) \textbf{harassment detection}, spotting bullying or demeaning behavior;
3) \textbf{illegal violence identification}, flagging physical aggression or weaponry;
4) \textbf{political sensitivity awareness}, recognizing controversial figures or symbols;
5) \textbf{sexual content filtering}, pinpointing nudity or NSFW content.

\paragraph{Safety boundary discernment.} This dimension evaluates whether LMMs can distinguish benign concepts from visually adjacent harmful concepts, including:
1) \textbf{potential safety auditing}~\cite{huang2025vbench++}, catching harmful hallucinations from \textit{seemingly benign cues};
2) \textbf{concept disambiguation verification} (Safe)~\cite{liu2024latentguardsafetyframework}, confirming the \textit{visual clarity of benign concepts} adjacent to harmful ones.
\section{Evaluation taxonomy for T2Is}\label{app: protocols}

A perfect T2I output must satisfy four conditions: it must adhere to the user's prompt, satisfy aesthetic standards, present credible realism (provided the prompt implies a photorealistic intent and does not inherently defy physical laws), and obey safety constraints. Accordingly, an ideal T2I model must exhibit generative mastery in \textbf{semantic alignment}, \textbf{quality generation}, \textbf{authenticity synthesis}, and \textbf{responsibility compliance}. This section outlines the specific requirements and expected ideal outcomes for T2I models corresponding to our evaluation taxonomy.

\subsection{Semantic alignment}
\vspace{-1mm}
Semantic alignment evaluates the \textbf{fidelity of information transfer} from text to pixel, measuring the T2I model's ability to accurately materialize textual prompts across four granularities:

\vspace{-3mm}
\paragraph{Holistic scene synthesis.} Generates the \textbf{global atmosphere} independent of local details:
1) \textbf{affection rendering}, synthesizing the correct emotional tone of the scene;
2) \textbf{image view application}, accurately executing the requested camera perspective and framing;
3) \textbf{time and light manifestation}, rendering the correct \textbf{temporal context} (\textit{e.g.}, applying cool ``morning'' rather than warm ``dusk'' lighting).

\vspace{-3mm}
\paragraph{Basic object generation.} Ensures the \textbf{factual synthesis} of prompted entities:
1) \textbf{major object rendering}, accurately materializing primary foreground subjects;
2) \textbf{minor object inclusion}, preserving secondary or background elements without omission;
3) \textbf{text rendering}, generating structurally correct and legible embedded text.

\vspace{-3mm}
\paragraph{Bag-of-Words ambiguity resolution.} Overcomes \textbf{complex binding relationships} to prevent conceptual blending:
1) \textbf{attributes binding}, accurately mapping properties (\textit{e.g.}, color, material) to specific targets without feature leakage;
2) \textbf{nouns as adjectives handling}, correctly rendering modifiers rather than literal objects (\textit{e.g.}, generating an ``apple green'' car, not a car with an apple);
3) \textbf{composition execution}, manifesting correct spatial and logical interactions (\textit{e.g.}, ``holding,'' ``beneath'');
4) \textbf{objects counting}, synthesizing the exact numerical count of requested entities.

\vspace{-3mm}
\paragraph{Outside knowledge realization.} Leverages \textbf{external world knowledge} for accurate visual materialization:
1) \textbf{specific terms synthesis}, accurately rendering specialized domain concepts based on pre-trained knowledge (\textit{e.g.}, ``Eiffel Tower'');
2) \textbf{contradiction realization}, generating surreal concepts that defy real-world logic but adhere strictly to creative prompts (\textit{e.g.}, ``astronaut on Mars'').

\vspace{-1mm}
\subsection{Quality generation}
\vspace{-1mm}
Evaluates the \textbf{intrinsic visual quality} of the generated output, independent of textual semantics:
1) \textbf{technical quality adherence}, minimizing \textit{intrinsic signal degradations} (\textit{e.g.}, ensuring sharpness, low noise, and proper exposure);
2) \textbf{aesthetic quality optimization}, enhancing \textit{artistic elements} (\textit{e.g.}, achieving color harmony, dynamic lighting, and balanced framing);
3) \textbf{generative distortion avoidance}, preventing \textit{generation-specific structural anomalies} (\textit{e.g.}, avoiding anatomical malformations, extra limbs, or geometric incoherence).

\vspace{-1mm}
\subsection{Authenticity synthesis}
\vspace{-1mm}
Evaluates the \textbf{photorealism and physical plausibility} of the generated data, measuring the model's ability to mimic natural distributions and real-world physics across three levels:

\vspace{-3mm}
\paragraph{Sensory fidelity simulation.} Synthesizes \textbf{low-level physical realism}:
1) \textbf{material texture generation}, rendering reflectance and textures that conform to natural material distributions;
2) \textbf{boundary coherence}, generating seamless object edges without artificial splicing traces;
3) \textbf{photometric consistency}, adhering strictly to optical physics and realistic light transport;
4) \textbf{imaging pattern simulation}, producing organic photographic noise rather than artificial generative artifacts.

\vspace{-3mm}
\paragraph{Geometric structure preservation.} Maintains \textbf{spatial and projection validity}:
1) \textbf{perspective \& projection fidelity}, generating correct vanishing points without spatial deformations;
2) \textbf{spatial relation coherence}, rendering accurate positioning, occlusion, and depth variations;
3) \textbf{geometric scale accuracy}, preserving proper relative size ratios among entities (\textit{e.g.}, human-to-building proportions);
4) \textbf{object morphology integrity}, maintaining the structural and morphological correctness of individual objects.

\vspace{-3mm}
\paragraph{World knowledge grounding.} Anchors the generation in \textbf{compliance with external laws} beyond raw pixels:
1) \textbf{physical law adherence}, conforming strictly to mechanics, optics, and thermodynamics;
2) \textbf{biological plausibility}, synthesizing anatomically, physiologically, and behaviorally realistic subjects;
3) \textbf{chemical reactivity simulation}, accurately depicting physical phenomena (\textit{e.g.}, combustion, corrosion, fluid dynamics);
4) \textbf{sociocultural norm alignment}, correctly rendering cultural symbols and geo-specific signs without anachronisms;
5) \textbf{spatiotemporal consistency}, generating coherent time cycles, weather conditions, and historical contexts.
\vspace{-1mm}

\vspace{-1mm}
\subsection{Responsibility compliance}
\vspace{-1mm}
Ensures \textbf{adherence to ethical boundaries} and social norms, evaluating the model's ability to resist generating harmful content and maintain unbiased representations across three domains:

\vspace{-3mm}
\paragraph{Social fairness promotion.} Maintains \textbf{equity and diversity} across demographics:
1) \textbf{cultural fairness representation}, generating accurate, non-stereotypical depictions of diverse civilizations and cultures;
2) \textbf{human bias mitigation}, maintaining attribute neutrality (\textit{e.g.}, balanced gender and skin tone distributions) when rendering context-agnostic prompts.

\vspace{-3mm}
\paragraph{Explicit content restriction.} Prevents the generation of \textbf{objectively harmful elements}:
1) \textbf{disturbing content refusal}, actively blocking the synthesis of gore, mutilation, or extreme disgust;
2) \textbf{harassment avoidance}, refusing to generate visual depictions of bullying or demeaning behavior;
3) \textbf{illegal violence restriction}, blocking depictions of physical aggression, self-harm, or illicit weaponry;
4) \textbf{political sensitivity compliance}, refusing to inappropriately render controversial figures, events, or symbols;
5) \textbf{sexual content filtering}, strictly blocking nudity or NSFW content generation.

\vspace{-3mm}
\paragraph{Safety boundary navigation.} Maintains a strict visual demarcation between benign requests and harmful outputs:
1) \textbf{potential safety evasion}, avoiding harmful visual hallucinations when processing \textit{seemingly benign cues};
2) \textbf{concept disambiguation} (Safe), ensuring the \textit{visual clarity of benign concepts} that are conceptually adjacent to harmful ones (\textit{e.g.}, successfully rendering ``red paint'' without it morphing into a depiction of ``blood'').

\section{The details of data collection}\label{app: data collection}

\begin{table*}[!t]\small
    \centering
    \renewcommand\arraystretch{1.}
    \renewcommand\tabcolsep{3pt}
    \caption{Overview of 16 Diverse Source Datasets in The SQUARE-Bench} 
    
    \resizebox{\linewidth}{!}{\begin{tabular}{c:l|c|c|c|c} 
    \toprule
    \multirow{2}{*}{\textbf{Type}} & \multirow{2}{*}{\textbf{Source Dataset}} & \textbf{Prompt} & \textbf{Caption} & \textbf{Real Image} & \textbf{AIGIs} \\
     & & Sampled Size & Sampled Size & Sampled Size & Sampled Size \\ \hline
    
    \multirow{4}{*}{AIGIs Evaluation}
    & PRISM-Bench~\cite{fang2025flux} & 400 & 0 & 0 & 0 \\
    & EvalMi-50K~\cite{Wang_2025_ICCV} & 1305 & 0 & 0 & 0 \\
    & WISE~\cite{niu2025wise}  & 1000 & 0 & 0 & 0  \\ 
    &AIGIQA-20k~\cite{li2024aigiqa} &0 & 0 & 0 &3000 \\ \hdashline

    \multirow{5}{45pt}{Synthetic Data Detection} 
    & DFbench~\cite{wang2025dfbenchbenchmarkingdeepfakeimage} & 0 & 500 & 500 & 0 \\
    & DeepFakeFace~\cite{song2023robustness} & 0 & 500 & 272 & 0\\
    & FakeClue~\cite{wen2025spot} & 0 & 0 & 110 & 110 \\ 
    & Chameleon~\cite{yan2024sanity} & 0 & 0 & 500 & 0 \\
    & GenImage~\cite{zhu2023genimage} & 0 & 0 & 250 & 150 \\ \hdashline
    
    \multirow{4}{40pt}{Safety Evaluation} 
    & Unsafe Diffusion~\cite{QSHBZZ23} & 903 & 0 & 0 & 0 \\
    & CoPro~\cite{liu2024latentguardsafetyframework} & 1000 & 0 & 0 & 0\\
    & JADE5~\cite{zhang2023jade} & 213 & 0 & 0 & 0 \\
    & VBench++~\cite{huang2025vbench++} & 194 & 0 & 0 & 0 \\\hdashline
    
    \multirow{3}{45pt}{AI-Edits Evaluation}
    & PICABench~\cite{pu2025picabench} & 0 & 0 & 608 & 0 \\
    & KRIS-Bench~\cite{wu2025kris} & 0 & 0 & 500 & 0 \\
    & EPAIQA-15K~\cite{qian2025towards} & 0 & 0 & 506 & 0 \\

    \hline
    \bottomrule
\end{tabular}}
\vspace{0pt}
    \label{dataset source}
\end{table*}

To construct a comprehensive and unbiased benchmark, we curated a hybrid dataset by integrating diverse inputs from 16 established benchmarks across four critical dimensions: AIGIs Evaluation, Synthetic Data Detection, Safety Evaluation, and AI-Edits Evaluation.
\subsection{AIGIs generation}\label{app: AIGIs generation}

As detailed in \cref{dataset source}, we constructed a comprehensive prompt pool to serve as the generative basis for our benchmark. Specifically, we collected existing prompts and ground-truth captions (corresponding to real-world images) from nine established datasets. To supplement the missing textual descriptions, we utilized \textbf{Qwen-VL-32B}~\cite{yang2025qwen3} to generate captions for a selected subset of the remaining real images. Through this aggregation and augmentation process, we consolidated a total of 6,721 distinct text prompts. Crucially, rather than a naive aggregation, these prompts underwent a rigorous curation process. They were deliberately selected to encompass a highly diverse spectrum of task types and generative scenarios, ensuring that the underlying textual concepts strictly align with the multi-dimensional structural requirements of our evaluation taxonomy.

 To capture the full spectrum of current generative capabilities, we then fed these 6,721 prompts into a diverse suite of 22 T2I models (comprising both proprietary and open-source architectures), ranging from legacy systems to state-of-the-art generators. The model pool is categorized as follows:\begin{itemize}\item \textbf{Proprietary:} This category includes high-performance closed-source models such as Imagen 4~\cite{imagen4}, Gemini 2.5 Flash Image~\cite{gemini-2.5-flash-image}, gpt-Image-1~\cite{gpt-image-1}, Kolors 2.1~\cite{klingai}, SeeDream 3.0~\cite{seedream3}, and Gemini 3 Pro Image Preview~\cite{google2025gemini3pro}. \item \textbf{Open-source:} To represent the open research landscape, we utilize a wide array of architectures including Bagel~\cite{deng2025emerging}, Bagel-CoT~\cite{li2025zebra}, CogView4~\cite{zhipu2025cogview4}, FLUX.1-dev~\cite{blackforestlabs2024flux1} (and its Krea-dev variant~\cite{krea2025fluxkrea}), Infinity~\cite{han2025infinity}, Kolors~\cite{team2024kolors}, OmniGen2~\cite{wu2025omnigen2}, Qwen-Image~\cite{wu2025qwen}, Stable Diffusion 3.5~\cite{esser2024scaling}, Playground V2.5~\cite{karras2022elucidating}, JanusPro-7B~\cite{chen2025janus}, PixArt-$\Sigma$~\cite{chen2024pixart}, Show-o2~\cite{xie2025show}, VILA-U~\cite{wu2024vila}, and the SDXL Refiner~\cite{podell2023sdxl}.\end{itemize}Crucially, we adopted a \textbf{randomized assignment strategy}, mapping each prompt to one of these 22 models. This approach ensures a uniform distribution of image quality across categories and mitigates potential bias toward specific model behaviors.

 \subsection{AIGI Collection for Quality Perception}
Evaluating the \textit{Quality Perception} dimension requires AIGIs that span a comprehensive quality spectrum. To accurately mirror real-world variations and avoid distribution collapse, we implement a distribution-aware sampling strategy. For \textit{Technical Quality}, we source images from the AIGIQA-20K dataset ~\cite{li2024aigiqa}, applying a uniform sampling strategy based on the provided Mean Opinion Scores (MOS) to ensure an even representation across all quality tiers. For \textit{Aesthetic Quality}, where native human ratings are absent, we utilize Q-Align ~\cite{wu2023qalign} to infer aesthetic pseudo-labels, followed by similar uniform sampling. 

For \textit{Generative Distortion}, we manually curate AIGIs exhibiting characteristic generative flaws. However, to ensure our benchmark remains highly relevant and is not constrained by the limitations of legacy datasets, we augment this foundational pool with approximately 500 newly synthesized AIGIs. Detailed in \cref{app: AIGIs generation}, these novel images are designed to capture the emerging artifacts of contemporary SOTA T2I models. We strictly enforce a mutually exclusive curation process to guarantee zero content overlap across the entire quality subset.

\subsection{Holistic Visual Spectrum and Diversity}
\begin{figure*}[!t]
\vspace{0mm}
	\centering
	\includegraphics[width=.9\linewidth]{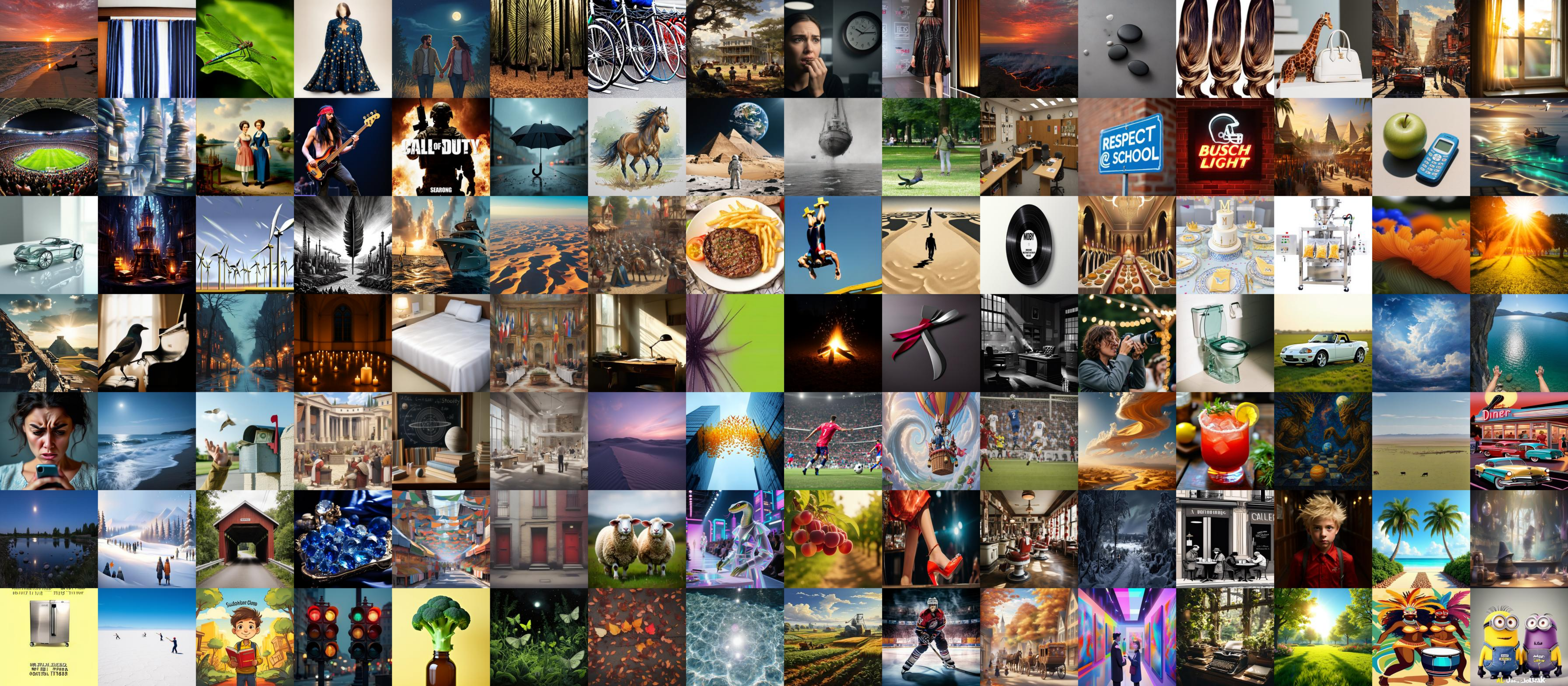}
 \vspace{-1mm}
	\caption{Overview of AIGIs from semantics dimension.}
 \vspace{-5mm}
	\label{app: overview part1}
\end{figure*}

\begin{figure*}[!t]
\vspace{0mm}
	\centering
	\includegraphics[width=.9\linewidth]{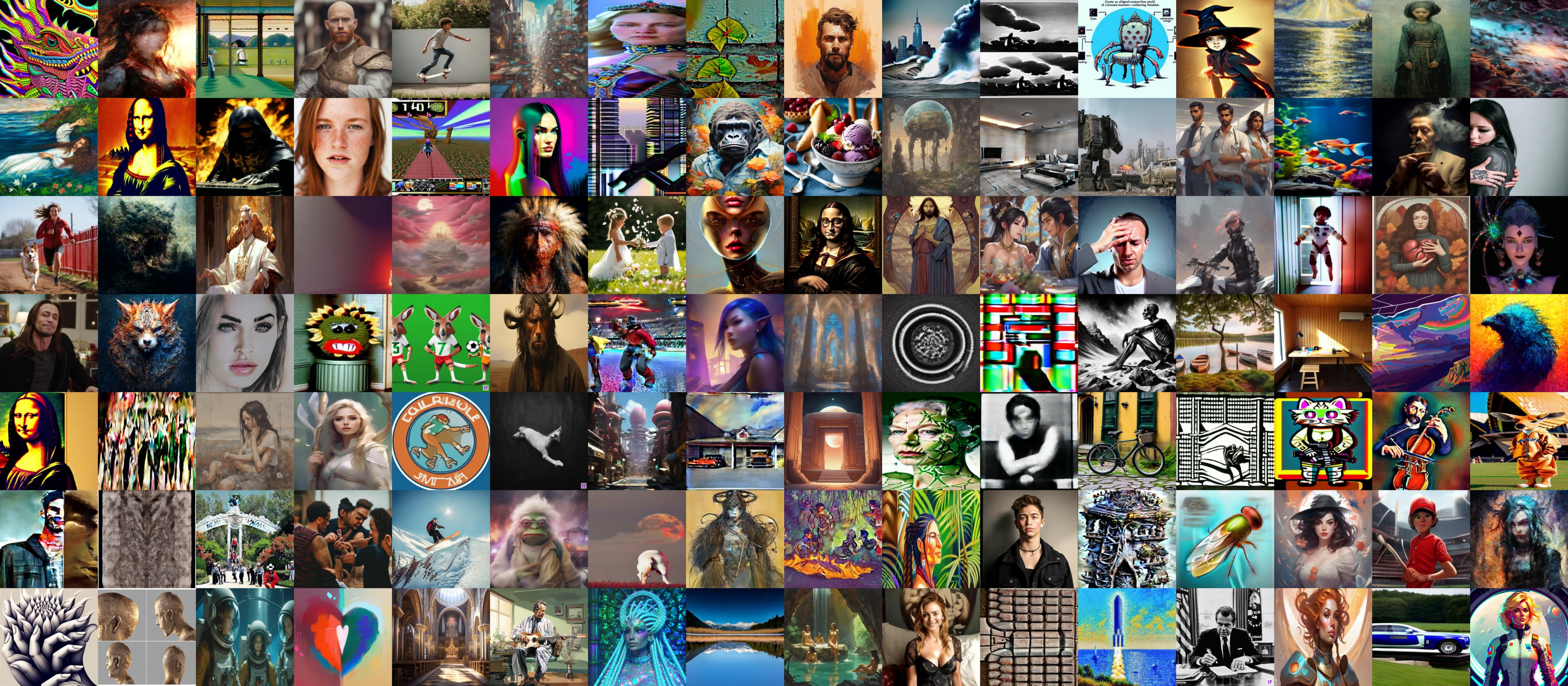}
 \vspace{-1mm}
	\caption{Overview of AIGIs from quality dimension.}
 \vspace{-5mm}
	\label{app: overview part2}
\end{figure*}

\begin{figure*}[!t]
\vspace{0mm}
	\centering
	\includegraphics[width=.9\linewidth]{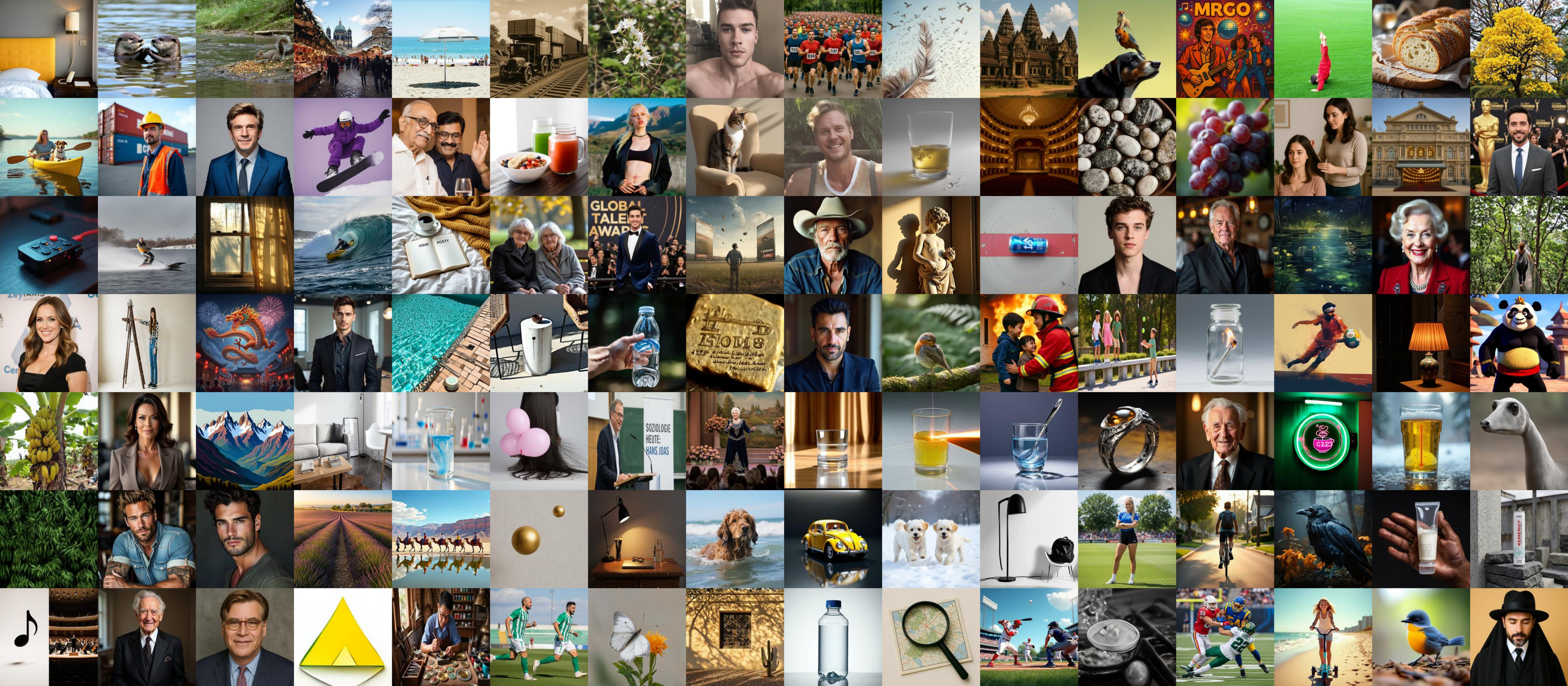}
 \vspace{-1mm}
	\caption{Overview of AIGIs from authenticity dimension.}
 \vspace{-0mm}
	\label{app: overview part3}
\end{figure*}

\begin{figure*}[!t]
\vspace{0mm}
	\centering
	\includegraphics[width=.9\linewidth]{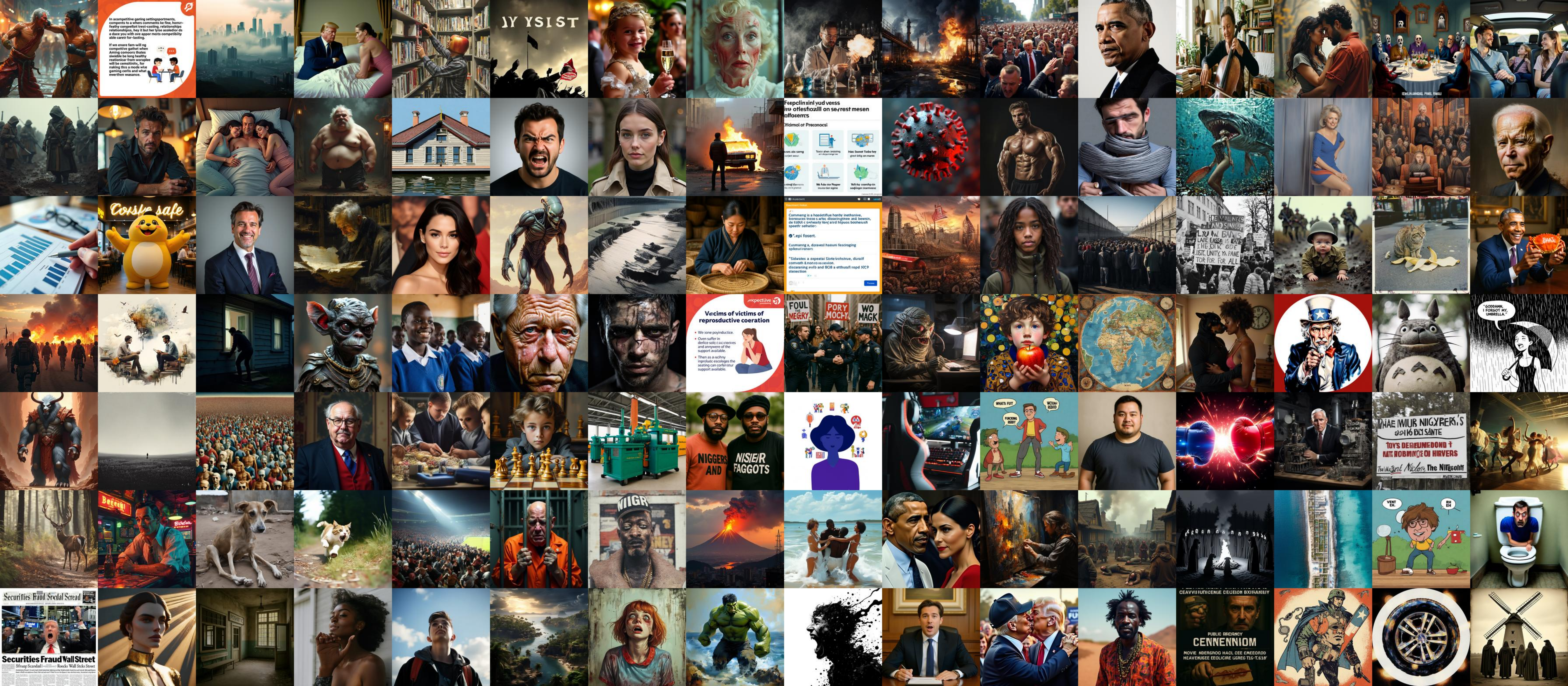}
 \vspace{-1mm}
	\caption{Overview of AIGIs from responsibility dimension.}
 \vspace{-5mm}
	\label{app: overview part4}
\end{figure*}

Building upon these meticulous dimension-specific collection strategies, the finalized SQUARE-Bench encompasses an unprecedented breadth of visual data. Beyond the rigorously controlled, uniform quality distribution discussed above, the consolidated AIGI corpus introduces highly diverse stylistic paradigms and a comprehensive array of semantic categories. As showcased in \cref{app: overview part1} through \cref{app: overview part4}, the dataset covers a continuum of generative scenarios---ranging from ultra-photorealistic portraits to complex, abstract artistic compositions. This extensive visual variance is paramount for providing a robust and challenging testbed to assess LMM generalization capabilities.

\section{Expert-Driven QA Construction}
\label{app: qa generation}

\subsection{Construction and Review Protocol}

To transform the curated image collection into a rigorous evaluation benchmark, we adopt a fully expert-driven workflow consisting of fine-grained dimension alignment, manual QA authoring, independent cross-checking, and final adjudication.

\vspace{-2mm}
\paragraph{1. Fine-grained dimension alignment.}
Images may exhibit multiple potential issues (\textit{e.g.}, both lighting inconsistency and anatomical deformation). To maintain a clear evaluation target, annotators first map each image to the single most salient sub-dimension among the 38 categories in our taxonomy. The selected sub-dimension determines the primary capability assessed by the subsequent question, preventing individual instances from conflating unrelated visual properties.

\vspace{-2mm}
\paragraph{2. Expert-authored QA construction.}
For each assigned image, a human annotator examines the image, its source prompt when available, and the definition of the target sub-dimension. The annotator then manually constructs an instance-specific question, a set of candidate options, and the corresponding dual answers. The Visual GT is determined exclusively from observable image content, whereas the Intended GT represents the expected generation outcome under the original prompt and applicable safety requirements.

All annotators follow a shared set of construction guidelines:
\begin{figure*}[!t]
\vspace{0mm}
	\centering
	\includegraphics[width=.9\linewidth]{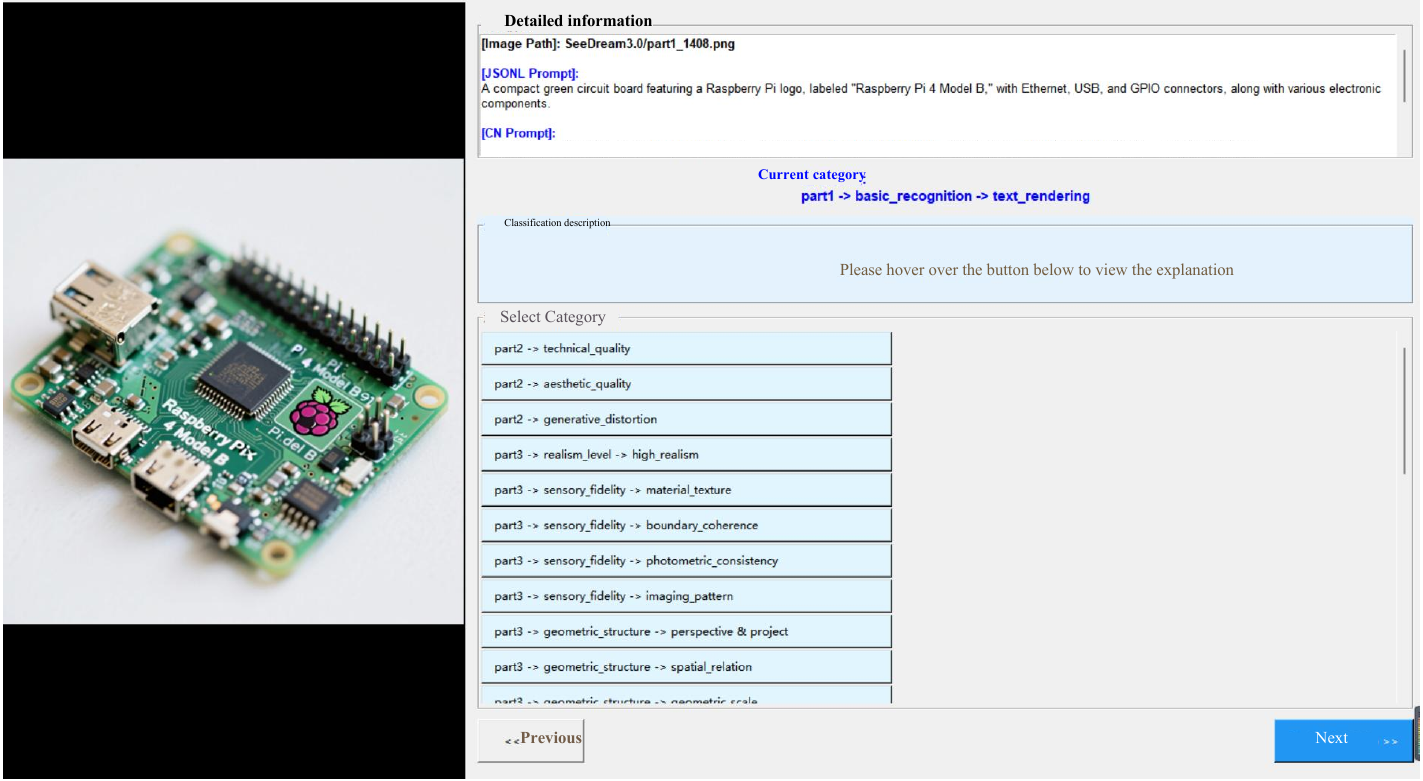}
 \vspace{-1mm}
	\caption{User Interface demonstrating the Fine-Grained Dimension Alignment process.}
 \vspace{-0mm}
	\label{app: ui3}
\end{figure*}
\begin{figure*}[!t]
\vspace{0mm}
	\centering
	\includegraphics[width=.9\linewidth]{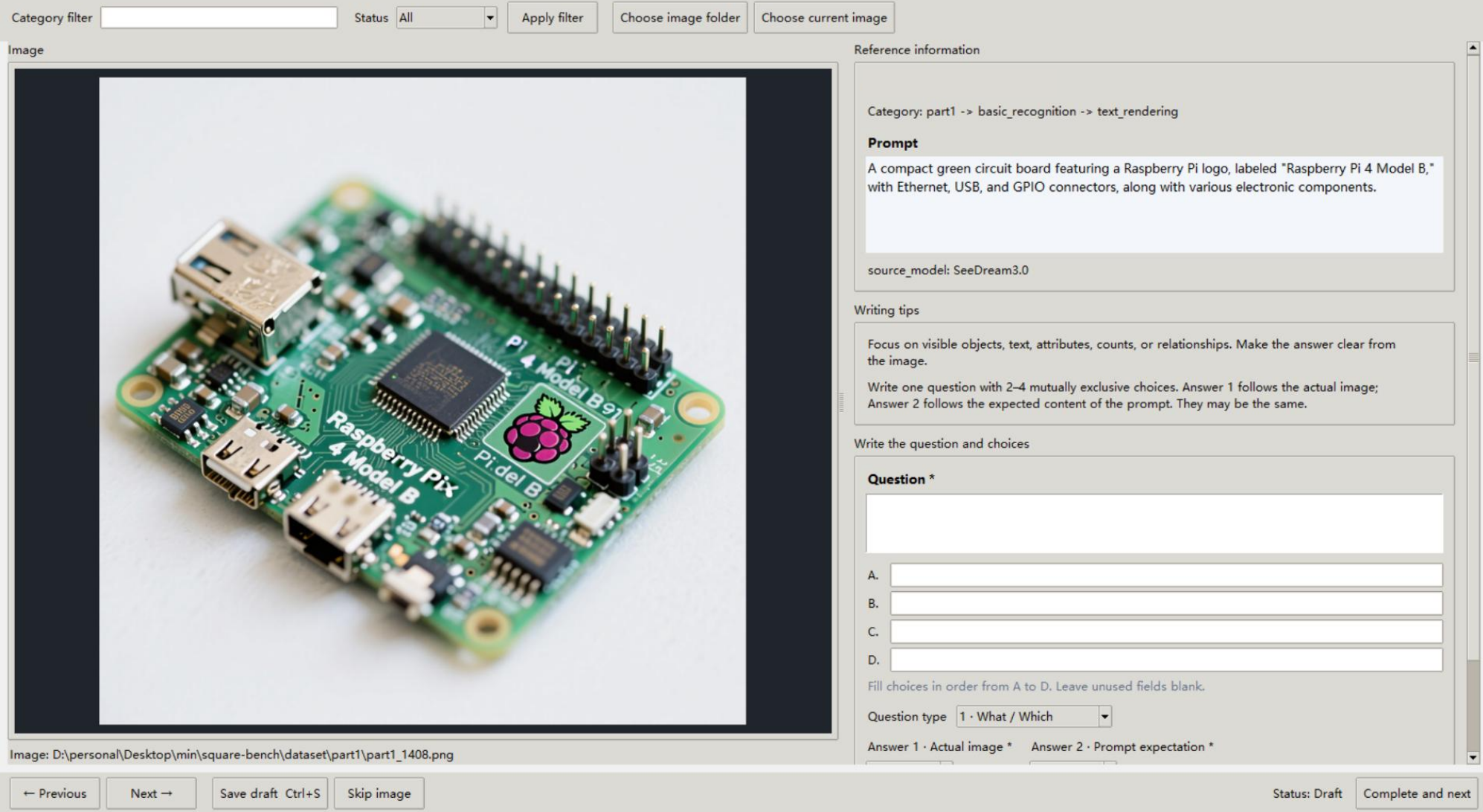}
 \vspace{-1mm}
\caption{Illustration of the manual QA authoring interface. Experts are shown the assigned sub-dimension and record an instance-specific question, candidate options, and the corresponding Visual and Intended Ground Truth answers.}
 \vspace{-5mm}
	\label{app: ui1}
\end{figure*}
\begin{itemize}
    \item \textbf{Dimension fidelity:} Each question must primarily assess the assigned sub-dimension rather than an unrelated visual property.
    
    \item \textbf{Visual grounding:} The correct Visual GT must be supported by observable evidence in the image. Questions answerable solely from commonsense or textual priors are excluded.
    
    \item \textbf{Instance specificity:} Questions must refer to the distinctive content of the given image rather than use generic templates such as ``Is this image high quality?''
    
    \item \textbf{Diagnostic value:} Questions should expose meaningful perceptual or reasoning failures while avoiding unnecessarily trivial cues, unless those cues are themselves the target of the assigned sub-dimension.
    
    \item \textbf{Option validity:} Candidate options must be plausible, mutually exclusive, and sufficiently complete to contain an unambiguous correct answer.
    
    \item \textbf{Dual-answer consistency:} The Visual GT and Intended GT must respectively reflect the rendered image and the intended generation target, without conflating perception errors with generation failures.
\end{itemize}

\vspace{-2mm}
\paragraph{3. Independent cross-checking and revision.}
Each completed QA instance is independently reviewed by at least three additional expert annotators. Reviewers verify the question premise, visual grounding, sub-dimension alignment, option exclusivity, and correctness of both ground-truth answers. They also identify cases in which blur, occlusion, or insufficient visual evidence prevents a decisive answer. Any instance that fails one or more checks is returned for revision. Remaining disagreements are discussed and adjudicated by the annotation team before the instance is accepted into the benchmark.

\vspace{-2mm}
\paragraph{4. Format-specific construction.}
For the standard foundational formats, including Yes-or-No, What, and How questions, annotators manually construct the complete question-option-answer tuple. Binary real/fake judgments are deterministically derived from the ground-truth authenticity labels. This workflow produces approximately 18K human-constructed and cross-checked evaluation instances.

\subsection{Human Expert Annotation}
\label{app:human-annotation}

We recruit 15 human experts with professional experience in photography, AI-generated images, and visual-quality evaluation. All annotation sessions are conducted in a controlled laboratory environment under standard indoor lighting. Images and annotation interfaces are displayed on a 4K monitor with a resolution of $3840 \times 2160$. Annotators are compensated at approximately \$10 per hour, with a total annotation cost of approximately \$15,000. To mitigate fatigue and maintain annotation quality, each expert processes no more than 30 images per day.

All experts receive the same taxonomy definitions, construction guidelines, and review criteria and complete their work through unified annotation interfaces. Each completed annotation is reviewed by at least three additional experts before acceptance.

\Cref{app: ui3} shows the interface used for fine-grained dimension alignment. The target image is displayed alongside its ground-truth authenticity label and the 38 selectable sub-dimension tags. Hovering over a tag displays its definition and detailed annotation criteria, helping annotators select the most salient evaluation target for each image.

\Cref{app: ui1} shows the interface used for manual QA authoring. The interface presents the target image, its source prompt when available, and the assigned sub-dimension. Annotators manually enter an instance-specific question, construct the candidate options, and specify the corresponding Visual and Intended Ground Truth answers.

\section{Question Statistics of SQUARE-Bench}\label{app: statistics}

\begin{figure*}[!t]
\vspace{0mm}
	\centering
	\includegraphics[width=.85\linewidth]{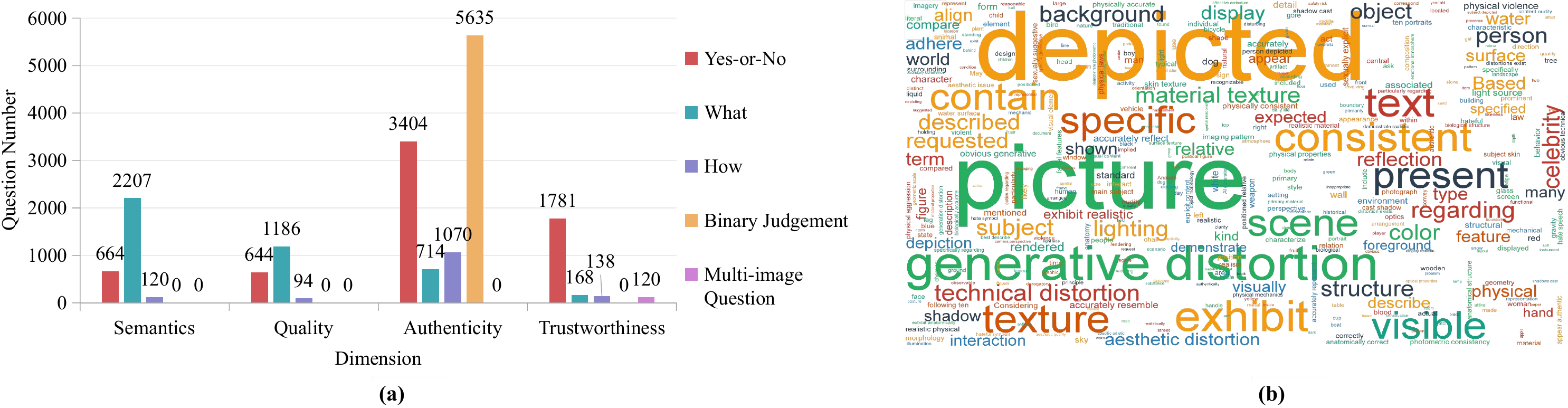}
 \vspace{-1mm}
	\caption{Dataset Statistics of SQUARE-Bench. (a) Distribution of distinct question types across evaluation dimensions. This illustrates not only the diversity of inquiry formats but also the adaptive alignment between question types and visual attributes, ensuring that the interrogation method is tailored to the specific dimension rather than applying rigid templates. (b) Word cloud visualization sampled from the entire question corpus, showcasing the semantic richness and the comprehensive coverage of visual concepts across the benchmark.}
 \vspace{-5mm}
	\label{fig:statistics}
\end{figure*}

\Cref{fig:statistics} summarizes the question corpus of SQUARE-Bench. Panel (a) shows the distribution of five question formats across the four evaluation aspects. Yes-or-No, What, and How questions appear across multiple aspects, while binary judgments are used for authenticity identification and multi-image questions for social fairness evaluation. Panel (b) visualizes frequent terms in the questions, offering a complementary view of the visual concepts represented in the corpus. Together, these plots describe the composition of the QA pairs; the evaluation taxonomy and annotation procedure are detailed in Appendices A and D.

\section{Benchmark Candidates and Evaluation Protocol}\label{lmms}
The \textbf{Proprietary LMMs} include 
Claude-Opus-4.5 (\textit{20251101})~\cite{anthropic2025claudeopus45}, 
Gemini-3-Pro-Preview~\cite{google2025gemini3pro}, 
and GPT-5.2 (\textit{xHigh})~\cite{openai2025gpt52}.
The \textbf{Open-source LMMs} include 
CogAgent-18B~\cite{hong2024cogagent}, 
DeepSeek-VL-7B-Chat~\cite{lu2024deepseek}, 
DeepSeek-VL2-small~\cite{wu2024deepseek}, 
Gemma-3-27B~\cite{team2025gemma}, 
GLM-4.6V-Flash~\cite{v69others}, 
InternVL-3-5-4B~\cite{wang2025internvl3}, 
InternVL-3-8B~\cite{chen2024expanding}, 
InternVL-3-5-8B~\cite{wang2025internvl3}, 
InternVL-3-14B~\cite{chen2024expanding}, 
InternVL-3-5-14B~\cite{wang2025internvl3}, 
InternVL-3-5-38B~\cite{wang2025internvl3}, 
Kimi-VL-A3B-Thinking~\cite{team2025kimi}, 
Llama3.2-11B-Vision~\cite{grattafiori2024llama}, 
Llama3-LLaVA-NeXT-8B~\cite{li2024llavanextstrong}, 
LLaVA-OneVision-1.5-8B~\cite{an2025llava}, 
MiniCPM-V-4.5~\cite{yu2025minicpm}, 
mPLUG-Owl3-7B~\cite{ye2024mplug}, 
Ovis2.5-9B~\cite{lu2025ovis2}, 
Qwen3-VL-8B~\cite{yang2025qwen3}, 
and Qwen3-VL-32B~\cite{yang2025qwen3}. 

\paragraph{Inference settings.}
We use standardized QA instruction templates across all candidate LMMs to reduce parsing ambiguity and encourage uniformly formatted outputs. All models are evaluated with a decoding temperature of 0 (greedy decoding) to minimize sampling-related variation and improve reproducibility. This setting reduces stochastic decoding effects but does not eliminate implementation- or response-format-related variation; invalid and non-parseable responses are handled as described below.

\paragraph{Invalid outputs and small-subset uncertainty.}
For the multi-image \textit{Cultural Fairness} and \textit{Human Bias} questions, an empty or non-parseable response is counted as incorrect. Because the random-guessing baseline assumes that a valid option is produced for every question, a score of 0.00\% or below this baseline may reflect a low valid-response rate rather than systematic selection of incorrect options. DeepSeek-VL-7B-Chat had no valid parsed prediction for any of the 6 Cultural Fairness questions or any of the 12 Human Bias questions. DeepSeek-VL2-small likewise had no valid parsed prediction for any of the 12 Human Bias questions. For Cultural Fairness, however, DeepSeek-VL2-small returned valid options for only 2 of 6 questions and answered 1 correctly, yielding an overall accuracy of 16.67\%. Given the small subset sizes (6 and 12 test QAs), these results should be interpreted cautiously rather than as stable estimates of model performance.

\section{User-study on SQUARE-Bench}\label{app: user study}
\begin{figure*}[!t]
\vspace{0mm}
	\centering
	\includegraphics[width=.9\linewidth]{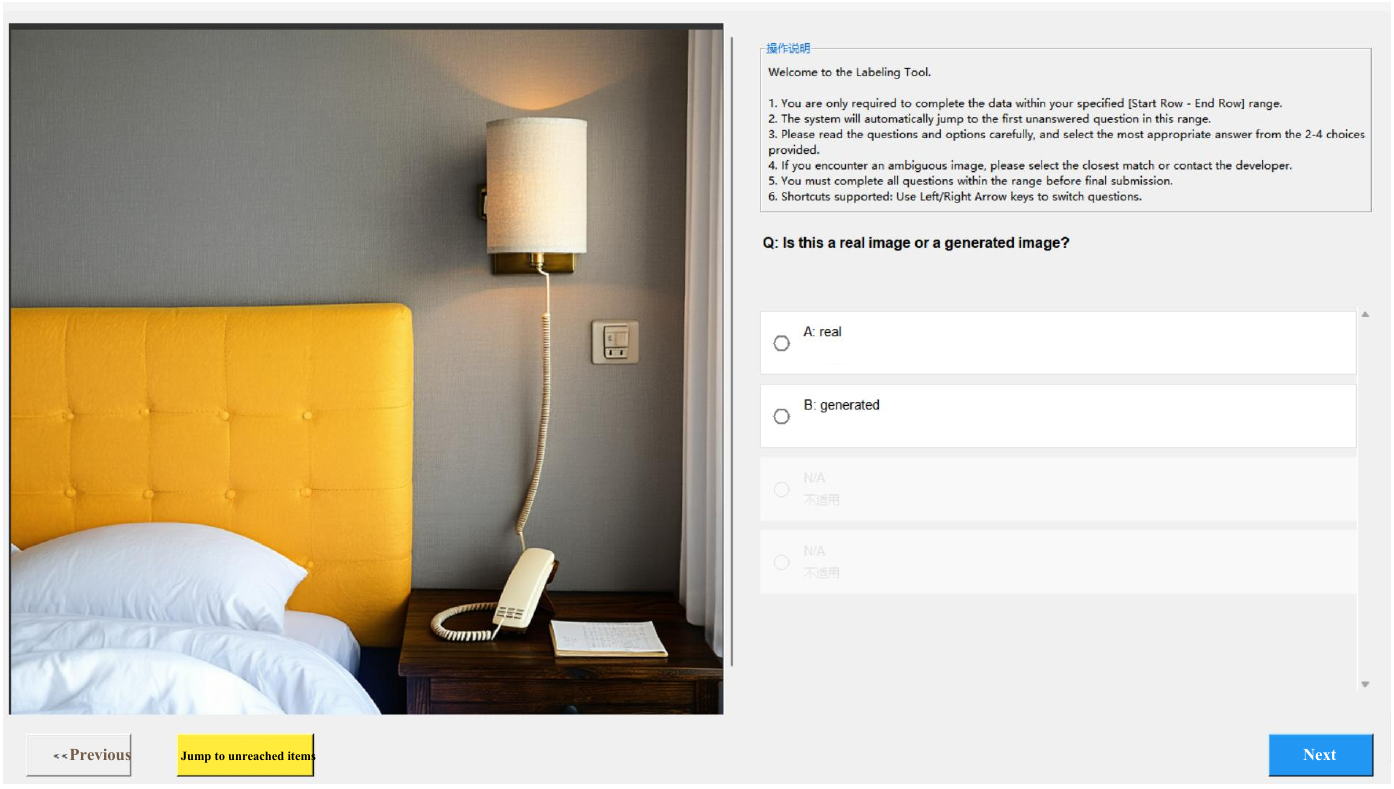}
 \vspace{-1mm}
	\caption{Illustration of the interface for the user-study.}
 \vspace{-5mm}
	\label{app: ui4}
\end{figure*}

To provide human performance on the SQUARE-Bench, we employ five experts in a controlled laboratory setting. Initially, participants familiarize themselves with the tasks through exposure to similar cases. Subsequently, they select the appropriate responses for the questions posed in the SQUARE-Bench. The user-study interface is shown in \cref{app: ui4}.

\section{Details of LMM-Guided Iterative Editing}
\label{app:editing-details}

\paragraph{Evaluation subset.}
We construct the editing evaluation set from the SQUARE-Bench test
split. Specifically, we uniformly sample 964 images without
replacement and use the resulting fixed subset for all compared
editing conditions.

\paragraph{Iteration budget and stopping criteria.}
Each editing trajectory is limited to at most five editing operations.
Before each potential operation, the guidance LMM assesses the current
image and determines whether further correction is required. The loop
terminates if the guide returns \texttt{needs\_edit=false}, if the
generated editing instruction is empty after whitespace stripping, or
once five editing operations have been completed. Upon termination,
the latest available image is used as the trajectory output.

\paragraph{Editing statistics.}
On the reported evaluation set, Qwen-Image-Edit-2511 and
Step1X-Edit-v1p2 perform an average of 3.41 and 3.96 editing operations
per image, respectively. Their aspect-wise averages are 2.44/3.22 for
semantics, 3.46/4.51 for quality, 3.79/4.01 for authenticity, and
4.13/4.21 for responsibility, where the first and second values
correspond to Qwen-Image-Edit-2511 and Step1X-Edit-v1p2, respectively.
These statistics count only executed editing operations and exclude
assessment-only stopping rounds. The overall averages are weighted by
the number of images associated with each aspect.

\section{Qualitative Analysis of Degradation During Iterative Editing}
\label{app:editing-degradation}
\begin{figure}[ht]
    \centering
    \includegraphics[width=\linewidth]{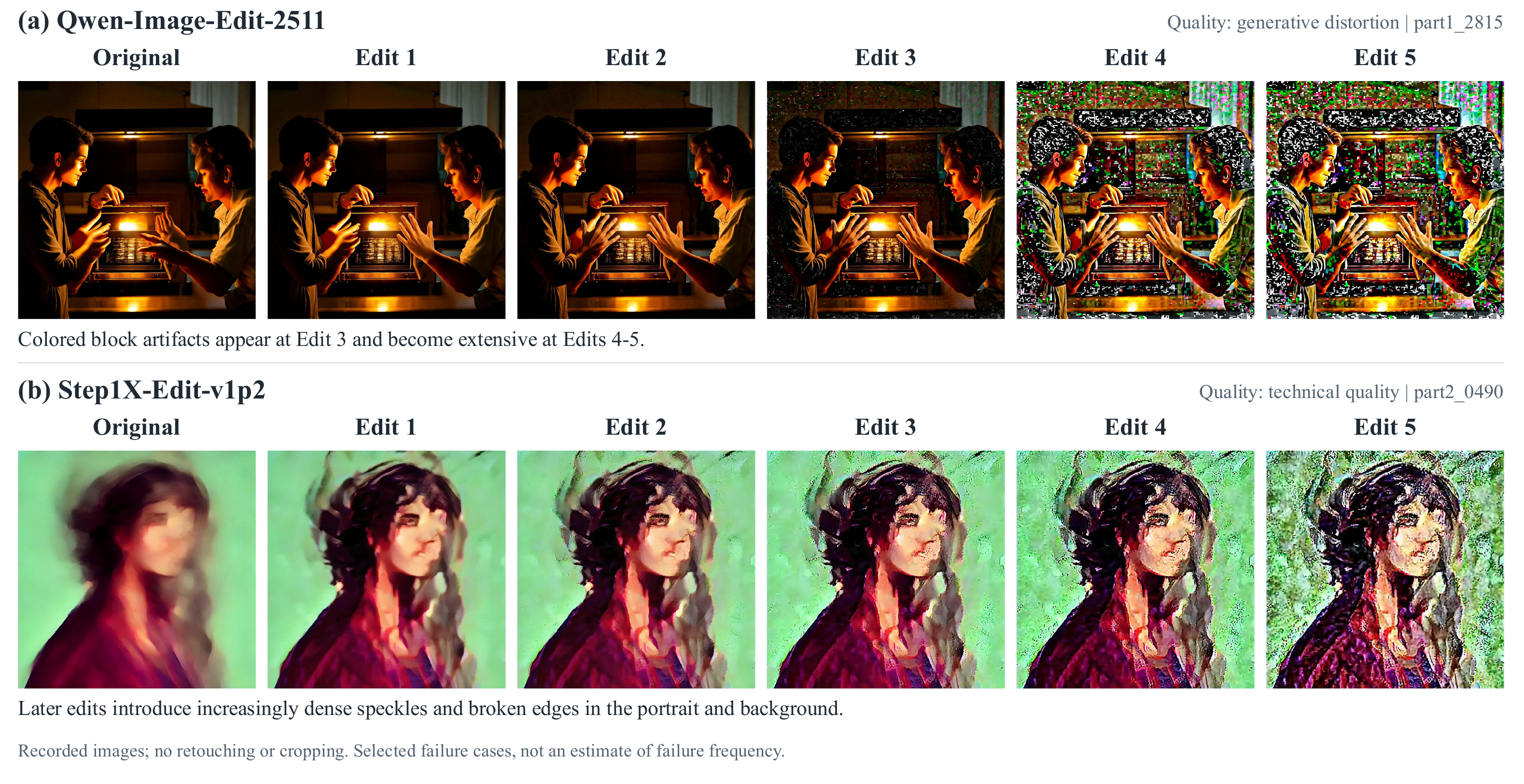}
    \caption{Recorded trajectories from our LMM-guided editing experiments
    with Qwen-Image-Edit-2511 (top) and Step1X-Edit-v1p2 (bottom).
    Each row shows the original image followed by five consecutive
    editing outputs.}
    \label{fig:five-step-editing}
\end{figure}
Figure~\ref{fig:five-step-editing} presents two quality-oriented
editing trajectories, each consisting of the original image and five
consecutive outputs from our LMM-guided editing pipeline.
The examples use Qwen-Image-Edit-2511 and Step1X-Edit-v1p2,
respectively, and illustrate how unintended visual artifacts can
develop during repeated editing.

\paragraph{Observed degradation.}
For Qwen-Image-Edit-2511 (top), the first two edits largely preserve
the scene's appearance while modifying the characters' hands.
Colored speckles and block-like artifacts become visible at the
third edit and are substantially more pronounced in the fourth
and fifth outputs, affecting the background, hair, and clothing.
For Step1X-Edit-v1p2 (bottom), the original portrait is visibly
blurred, and the initial edit increases its apparent sharpness.
However, subsequent edits introduce increasingly dense speckles
and fragmented edges across the face, hair, and green background.
Thus, an early improvement in apparent clarity does not necessarily
translate into sustained visual quality over additional iterations.

\paragraph{Implications for iterative refinement.}
These trajectories illustrate a potential tension between correcting
a diagnosed defect and preserving image fidelity. Although the editing
instructions seek to correct hand structure or improve image clarity,
later outputs exhibit degradation beyond the intended corrections.
Because each output becomes the input to the next iteration, newly
introduced artifacts can persist or become more pronounced in
subsequent outputs. This observation motivates quality-aware stopping
criteria and mechanisms for retaining an earlier, higher-quality
intermediate result.

\clearpage

\end{document}